\documentclass[10pt,journal,compsoc]{IEEEtran}
\ifCLASSOPTIONcompsoc
  \usepackage[nocompress]{cite}
\else
  \usepackage{cite}
\fi
\ifCLASSINFOpdf
\else
\fi

\usepackage[utf8]{inputenc} 
\usepackage[T1]{fontenc} 
\usepackage[]{hyperref} 
\usepackage{url} 
\usepackage{booktabs} 
\usepackage{amsfonts} 
\usepackage{nicefrac} 
\usepackage{microtype} 

\usepackage{graphicx}
\usepackage{subfigure}
\usepackage{times}
\usepackage{soul}
\usepackage{amsmath}
\usepackage{amsthm}
\usepackage{booktabs}
\usepackage{multirow}
\usepackage{algorithm}
\usepackage{algorithmic}
\usepackage{enumitem}
\usepackage{subfigure}
\usepackage{color}
\usepackage{bm}
\usepackage{cite}
\usepackage{orcidlink}
\usepackage{mathtools}

\usepackage{amsthm,amsmath,amssymb}
\usepackage{mathrsfs}

\newcommand{\Amat}[0]{\ensuremath{{\bf A}}}

\newcommand{\Hmat}[0]{\ensuremath{{\bf H}} }

\newcommand{\Kmat}[0]{\ensuremath{{\bf K}} }

\newcommand{\Mmat}[0]{\ensuremath{{\bf M}} }

\newcommand{\Smat}[0]{\ensuremath{{\bf S}} }

\newcommand{\Wmat}[0]{\ensuremath{{\bf W}} }
\newcommand{\Xmat}[0]{\ensuremath{{\bf X}} }

\newcommand{\Phimat}[0]{\ensuremath{{\bf \Phi}} }
\newcommand{\Lambdamat}[0]{\ensuremath{{\bf \Lambda}} }
\newcommand{\Omegamat}[0]{\ensuremath{{\bf \Omega}} }

\newcommand{\av}[0]{\ensuremath{\boldsymbol{a}} }

\newcommand{\hv}[0]{\ensuremath{\boldsymbol{h}} }

\newcommand{\kv}[0]{\ensuremath{\boldsymbol{k}} }

\newcommand{\uv}[0]{\ensuremath{\boldsymbol{u}} }

\newcommand{\xv}[0]{\ensuremath{\boldsymbol{x}} }

\newcommand{\phiv}[0]{\ensuremath{\boldsymbol{\phi}} }

\newcommand{\gammav}[0]{\ensuremath{\boldsymbol{\gamma}} }
\newcommand{\thetav}[0]{\ensuremath{\boldsymbol{\theta}} }
\newcommand{\lambdav}[0]{\ensuremath{\boldsymbol{\lambda}} }

\newcommand{\varepsilonv}[0]{\ensuremath{\boldsymbol{\varepsilon}} }

\newcommand{\given}{\mid}

\def\bb{}

\newcommand{\ns}{\negthickspace}

\begin{document}

\title{Scalable Weibull Graph Attention Autoencoder for Modeling Document Relational Networks}

\author{Chaojie~Wang*~\orcidlink{0000-0002-7644-7621},~\IEEEmembership{Member,~IEEE,}
        Xinyang~Liu*,
        Dongsheng Wang,
        Hao Zhang~\orcidlink{0000-0002-2928-2692}, \\
        Bo Chen~\orcidlink{0000-0001-5151-9388},~\IEEEmembership{Senior Member,~IEEE,}
        and Mingyuan Zhou
\thanks{$\bullet$ C. Wang, X. Liu, D. Wang, H. Zhang, and B. Chen are with School of Electronic Engineering, Xidian University, Xi'an, Shaanxi 710071, China (email: xd\_silly@163.com; lxy771258012@163.com; bchen@mail.xidian.edu.cn).}
\thanks{$\bullet$ M. Zhou is with McCombs School of Business, The University of Texas at Austin, TX 78712, USA (e-mail: mingyuan.zhou@mccombs.utexas.edu).}
\thanks{*The first two authors are equal contribution.}
\thanks{$\dagger$ Submit to T-PAMI}
}

%

\IEEEtitleabstractindextext{%

\begin{abstract}
Although existing variational graph autoencoders (VGAEs) have been widely used for modeling and generating graph-structured data, most of them are still not flexible enough to approximate the sparse and skewed latent node representations, especially those of document relational networks (DRNs) with discrete observations. To analyze a collection of interconnected documents, a typical branch of Bayesian models, specifically relational topic models (RTMs), has proven their efficacy in describing both link structures and document contents of DRNs, which motives us to incorporate RTMs with existing VGAEs to alleviate their potential issues when modeling the generation of DRNs. In this paper, moving beyond the sophisticated approximate assumptions of traditional RTMs, we develop a graph Poisson factor analysis (GPFA), which provides analytic conditional posteriors to improve the inference accuracy, and extend GPFA to a multi-stochastic-layer version named graph Poisson gamma belief network (GPGBN) to capture the hierarchical document relationships at multiple semantic levels. Then, taking GPGBN as the decoder, we combine it with various Weibull-based graph inference networks, resulting in two variants of Weibull graph auto-encoder (WGAE), equipped with model inference algorithms. Experimental results demonstrate that our models can extract high-quality hierarchical latent document representations and achieve promising performance on various graph analytic tasks.

\end{abstract}

}

\maketitle

\IEEEdisplaynontitleabstractindextext

%
\IEEEpeerreviewmaketitle

\IEEEraisesectionheading{\section{Introduction} \label{sec_intro}}
\IEEEPARstart{A} wide variety of network data, such as citation networks \cite{zhang2016graph}, chemical molecular structures \cite{liu2018constrained}, and recommendation systems \cite{wu2020diffnet++}, can be represented as a graph composed of a set of entities (nodes) and their links (edges).
Taking the graph characterized by node features and adjacency matrix as input, graph neural networks (GNNs) \cite{kipf2016semi, velickovic2018graph, GeCNs, TGNN} have demonstrated their efficacy in aggregating graph structural information and also achieved great success in graph signal processing \cite{DBLP:journals/tsp/ChenEZ21}.
\bb{However, for many practical graph analytic applications, especially for link prediction, modeling uncertainty \cite{kipf2016variational, hasanzadeh2019semi} in the latent space, rather than providing deterministic node representations, is of crucial importance, which can be potentially realized via introducing probabilistic generative models (PGMs) into existing GNNs.}

\bb{For graph analysis, various PGMs have been developed to discover the underlying relationships between nodes from their link structures via modeling the generative process of network data \cite{rematch}.
Specifically, the mechanism of variational autoencoder (VAE) \cite{kingma2014stochastic, rezende2014stochastic} has been successfully extended for modeling graph-structured data, resulting in a series of variational graph autoencoders (VGAEs) \cite{kipf2016variational} parameterized by GNNs \cite{kipf2016semi}.
Focused on task-specific network applications, several VGAE-based variations \cite{liu2018constrained, simonovsky2018graphvae, samanta2019nevae, zhang2019d} have been proposed for dealing with various practical graph analytic tasks and achieved promising experimental results.
However, the construction of existing VGAEs still heavily relies on the reparametrization of Gaussian-distributed latent node representations, which often fails to approximate the sparse and skewed ones \cite{zhang2018whai}, like the latent representations of document relational networks (DRNs) with discrete observations.
Although there have been some VGAE-liked methods \cite{hasanzadeh2019semi, davidson2018hyperspherical} attempting to learn non-Gaussian latent node representations to enrich the flexibility of posterior distributions, there remains a great challenge for these existing VGAEs to interpret the latent semantic structure learned from network data, especially the underlying topic connections between documents in a DRN, due to ignoring the reconstruction aspect of document contents.}

\bb{As another typical branch of PGM, Bayesian models have also shown their effectiveness in exploring and interpreting the latent structure of network data.
For instance, inspired by the success of applying topic models, with latent Dirichlet allocation (LDA) \cite{blei2003latent} being the best known representative, for non-negative matrix factorization, a series of relational topic models (RTMs) \cite{sachan2012using, chang2009relational,chang2010hierarchical, liu2009topic, sun2009itopicmodel, bavota2013methodbook} have been developed for describing the probabilistic generation process of DRNs, whose links are built based on document contents \cite{docenhance}.
Benefiting from jointly modeling the document contents represented as bag-of-words vectors and their relationships encapsulated as an adjacency matrix, these RTMs and their sophisticated variations \cite{zhang2012joint, he2015hawkestopic, guo2015Bayesian} have shown the promise on explainable link structure prediction, where each link can be interpreted with the common topic selection shared by the corresponding adjacent documents.
Despite achieving appealing performance, these existing RTMs still 
employ shallow model structures and under-exploit hierarchical semantics from document networks, which motivates us to construct a deep (hierarchical) RTM to capture multilevel semantic relationships based on the data augmentation technique adopted by recent popular deep probabilistic topic models \cite{ranganath2015deep, gan2015scalable, zhou2016augmentable}.
}

\bb{As mentioned above, the introduction of RTMs could be one of the most ideal choices to alleviate the potential issues when extending VGAEs for modeling DRNs. 
In this paper, to take both the advantages of RTMs and VGAEs while moving beyond their constraints, we first construct an interpretable deep RTM, and further extend it to a non-Gaussian VGAE with multiple stochastic hidden layers.
Besides DRNs, which has been widely used as a typical benchmark type in graph analytic tasks \cite{chen2022bag}, we emphasize that our developed models can be flexibly extended for dealing with other types of networks, whose node features and edges can be arbitrarily characterized as count, binary, or positive variables, with the link functions described in Appendix A.}
The main contributions of this paper have been summarized as follows:
\begin{itemize}[leftmargin=*]
	\setlength{\itemsep}{0pt}
	\setlength{\parsep}{0pt}
	\setlength{\parskip}{0pt}
    \item Based on the Poisson-gamma distribution, a novel RTM named graph Poisson factor analysis (GPFA) equipped with analytic conditional posteriors is developed, which jointly models both node features and link structure by sharing their latent representations (topic proportions).
    \item To explore hierarchical latent representations of nodes and reveal their underlying relationships at multiple semantic levels, we extend the developed GPFA to graph Poisson gamma belief network (GPGBN), which is the first unsupervised deep RTM for analyzing network data as far as we known.
    \item To achieve fast in out-of-sample prediction, we combine GPGBN (decoder) with a Weibull-based graph variational inference network (encoder), leading to a Weibull graph autoencoder (WGAE), which moves beyond traditional Gaussian-based VGAEs and can be flexibly generalized to handle with various graph analytic tasks.
    \item For efficient end-to-end model training, we propose both full-batch (non-scalable) and scalable versions of hybrid model inference algorithms for the developed WGAE, where the latter is scalable to large-scale graphs.
    \item Besides achieving state-of-the-art or comparable performance on various graph analytic tasks, our developed models provide a potential solution to exploring explainable link structure prediction at multiple semantic levels.
\end{itemize}

We note that this work is the journal extension of our conference paper presented in Wang et al. \cite{wang2020deep}, but there are several critical additional contributions in this paper to improve both model performance and efficiency, which can be roughly divided into two aspects as follows:
1) by introducing the Bayesian attention mechanism \cite{fan2020Bayesian} into the graph inference network, we develop a novel attention-based variant of WGAE, referred to as Weibull Graph Attention Autoencoder (WGAAE), which can leverage stochastic attention weights and significantly improve the model performance on a series of graph analytic tasks;
2) in order to reduce the heavy quadratic complexity caused by the graph reconstruction during training, 
we develop a novel subgraph decoding scheme for scalable model inference, which can effectively speed up the training procedure of WGAE under the premise of almost no loss of model performance, and the main idea can be flexibly extended to other existing VGAEs.

The remainder of this paper has been organized as follows:
In Section~\ref{sec_gpgbn}, we describe the technical details of the developed GPFA and extend it to the first deep RTM named GPGBN, which can explore hierarchical latent document representations and their underlying relationships on multiple semantic levels, followed by a discussion about model properties.
Section~\ref{sec_wgcae} introduces two different forms of Weibull-based graph inference network (encoder), one based on the vanilla GCN and the other based on the Bayesian attention mechanism, and combine them with GPGBN (decoder) into two specific variants of WGAE, which can be trained with either a full-batch (non-scalable) or a scalable model inference algorithm; 
Section~\ref{sec_experiments} evaluates the proposed methods with
both qualitative and quantitative experiments on various graph benchmarks.

\section{Related Works} \label{sec_related_work}

\subsection{Relational Topic Models}
Derived from traditional probabilistic topic models \cite{blei2003latent, canny2004gap}, RTMs \cite{chang2010hierarchical, liu2009topic} are developed to jointly model the generation of documents and their links built based on document contents.
Specifically, each document represented as a bag-of-word (BoW) vector exhibits a latent mixture of topics, and the connections between documents are modeled as binary variables depended on the topic assignments of the word tokens.
{Borrowing similar ideas from RTMs, a series of sophisticated probabilistic generative models (PGMs) \cite{zhang2012joint, he2015hawkestopic, guo2015Bayesian} have been developed for jointly modeling document networks with topic models, which can also be flexibly extended for the applications of other fields, such as social recommendation \cite{gopalan2013scalable, gopalan2014content}.}
However, limited by the non-conjugacy between the prior and the link likelihood, traditional RTMs usually utilize variational inference (VI) to perform posterior inference with the mean-field assumption, which has been proven too restrictive in practice.
To this end, making full use of the data augmentation techniques, 
more sophisticated RTMs  \cite{chen2014discriminative, acharya2015gamma, DATM, DUI} equipped with analytic posteriors are developed to be trained with the efficient collapsed Gibbs sampling, aimed at providing more accurate parameter estimations.

However, these mentioned RTMs are usually constrained by the single-layer network structure to sufficiently model the generation of both document contents and relationships, leading to their under-exploiting hierarchical semantics from DRNs.
To this end, with the data augmentation techniques \cite{ranganath2015deep, gan2015scalable, zhou2016augmentable}, we develop the first deep RTM with multiple stochastic layers to extract hierarchical latent document representations from DRNs and try to interpret their underlying relationships at multiple semantic levels, which has never been explored as far as we known.

\subsection{Graph Autoencoder}

Recently, graph neural networks (GNNs) \cite{wu2019comprehensive} have demonstrated their efficacy in aggregating information from graph-structured data, and become a powerful tool to explore the relational structure among objects.
Essentially, the edges in a graph can be treated as a special attention mechanism to indicate the strength of the relationship between two nodes (samples), and there has been recent interests to develop attention-based GNNs \cite{velickovic2018graph, wang2020learning}.
Further, by extending VAE \cite{kingma2014stochastic} for graph-structured data, VGAE \cite{kipf2016variational} is developed for learning graph representations in an unsupervised manner, which utilizes GCNs \cite{kipf2016semi} to encode both the node features and adjacency matrix with a Gaussian-parameterized inference network (encoder), and then reconstructs the observed adjacency matrix with a generative network (decoder).
To enrich the flexibility of posterior distributions, a series of VGAE-liked methods \cite{hasanzadeh2019semi, yin2018semi} has been developed to move beyond the naive choice of Gaussian-based latent representations to better approximate more complex posteriors.

Although having achieved promising performance on several  graph benchmarks especially document networks, limited by the design of uninterpretable generative process, there remains a great challenge for these existing VGAEs to interpret the latent semantic structures learned from network data, such as the hierarchical topics of document contents, and the multilevel underlying relationships between documents.
Furthermore, ignoring the reconstruction information of node features in these VGAEs can also harm their model performance on node clustering or classification, which motivates us to construct a novel hierarchical VGAE-liked model to take a deep RTM as the generative network (decoder).

\section{Deep relational topic models}\label{sec_gpgbn}

In the following discussion, we focus on modeling the widely used document relation networks (DRNs), which owns interpretable semantics, contributing to investigating the latent semantic structures learned from network data more intuitively.
Aimed at exploring the hierarchical semantic topics and underlying relationships among DRNs, in what follows, we will first discuss how to represent a DRN as a graph and then introduce the basic GPFA, which can be further extended to the first deep RTM named GPGBN.
For ease of understanding, we first attempt to model the generation of DRNs consisting of the most common count node features and binary links, which can be further generalized to other data types, as discussed in Appendix A.

\subsection{Representing a DRN as a Graph}

Defining $\mathcal{V}$ as the set of document nodes and $\mathcal{E}$ as the set of edges, we first formulate a DRN containing $N=\left| \mathcal{V} \right|$ documents as an undirected graph  $\mathcal{G} = \{\mathcal{V} ,\mathcal{E} \}$.
Following that, we will describe how to represent the node features (document representations) and edges (links between documents) in our models.

\textbf{Node Features:} Most DRNs consider the bag-of-words (BoW) vectors as document representations to capture the global semantics of documents, such as word co-occurrence patterns, inspiring us to represent the $i$-th document as a high-dimensional sparse count vector $\xv_i \in \mathbb{Z}^{K_0}$, where $\mathbb{Z}=\{ 0,1,...\}$ and $K_0$ is the vocabulary size.
Thus, the node features of $\mathcal{G}$ can be represented as a count matrix $\Xmat = [\xv_1, \cdots, \xv_N] \in \mathbb{Z}^{K_0 \times N}$.

\textbf{Adjacency Matrix:} The edges in a DRN can be encapsulated as an $N \times N$ adjacency matrix $\Amat= \{a_{ij}\}_{i=1,j=1}^{N,N}$, where $a_{ij}$ denotes the link between documents $i$ and $j$, taking binary, count, or real positive values \cite{cai2010graph, chang2010hierarchical}.
Since most existing DRNs employ a binary link to denote whether the two documents are connected, we will also focus on this common scenario in the following discussion.

\subsection{Graph Poisson Factor Analysis} \label{subsec_gpfa}

\begin{figure}[!t]
	\centering
	 \subfigure[GPFA]{\includegraphics[width=45mm]{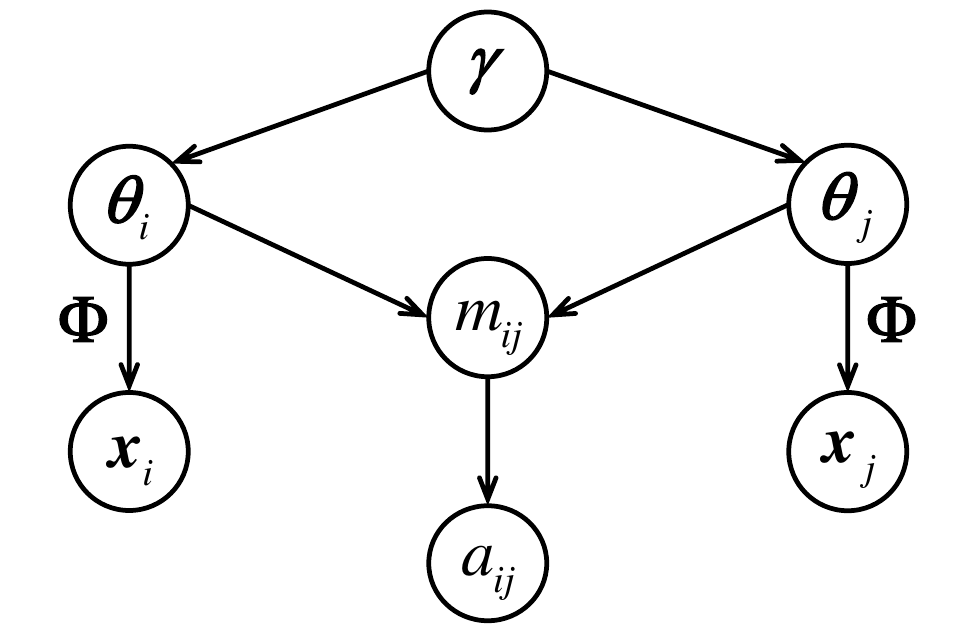}
		\label{fig_gpfa}} \quad \quad \quad
	 \subfigure[GPGBN]{\includegraphics[width=45mm]{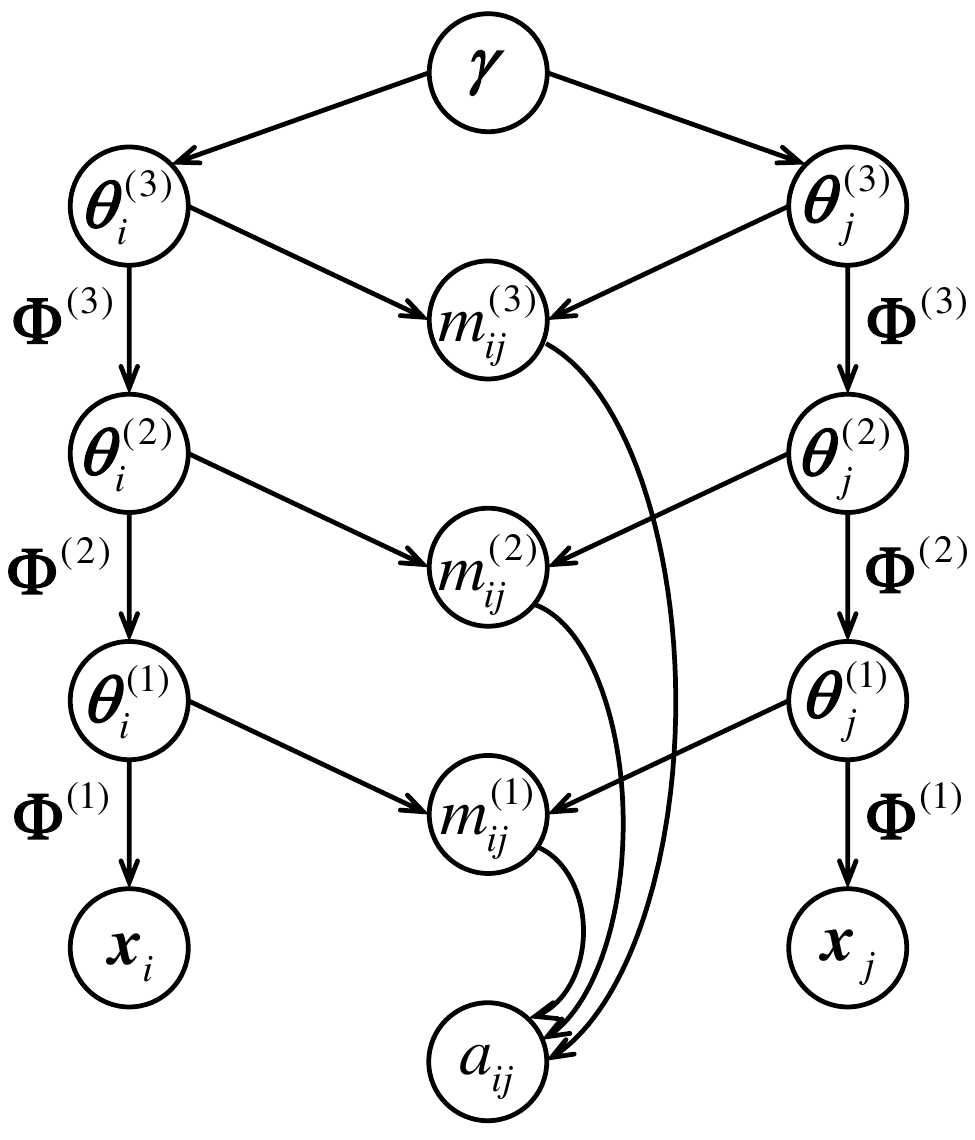}
		\label{fig_gpgbn}}
	\caption{The probabilistic generative process of (a) Graph Poisson Graph Factor Analysis (GPFA) and (b) Graph Poisson Gamma Belief Network (GPGBN).}
	\label{fig_generative_model}
\end{figure}

After representing the observed DRN as an undirected graph $\mathcal{G}$ consisting of the node features $\Xmat \in \mathbb{Z}^{K_0 \times N}$ and the adjacency matrix $\Amat \in \{0, 1\}^{N \times N}$, we follow the concept of designing RTMs \cite{chang2010hierarchical, liu2009topic} and develop a novel Graph Poisson Factor Analysis (GPFA) to model the generation of a DRN.
As illustrated in Fig.~\ref{fig_gpfa}, the developed GPFA can model the node-feature likelihood $p(\Xmat | \cdot)$ and edge likelihood $p(\Amat | \cdot)$ jointly, whose generative process can be formulated as 
\begin{align}\label{eq_GPFA}
&\thetav_j \sim \mbox{Gam}(\gammav, 1/c_j), \quad \phi_k \sim \mbox{Dir}(\eta), \notag\\
&m_{ij} \sim \mbox{Gam}(\sum\nolimits_{k = 1}^{K_1} {{u_k}{\theta _{ik}}{\theta _{jk}}}), \quad u_k \sim \mbox{Gam}({\alpha_k}, 1/\beta_k), \notag \\
&\xv_j \sim \mbox{Pois}(\Phimat~\thetav_j), \quad a_{ij} = {\bf{1}}_{m_{ij} \textgreater 0}, \end{align}
where the $j$th document represented as a BoW vector $\xv_j \in  \mathbb{Z}^{K_0}$ is generated from the product of the factor loading matrix (topics) $\Phimat =[\phiv_1,\ldots,\phiv_{K_1}]  \in \mathbb{R}_ + ^{{K_0} \times {K_1}}$ and the corresponding gamma distributed factor scores (unnormalized topic proportions) ${\thetav_j} = [\theta_{j1}; \ldots; \theta_{j K_1}] \in \mathbb{R}_ + ^{{K_1}}$ under the Poisson likelihood, with $K_1$ denoting the number of topics;
the binary edge $a_{ij}\in \{0, 1\}$ is generated by first drawing a latent count $m_{ij} \in \mathbb{Z}$ from the Poisson distribution with rate parameter $\sum\nolimits_{k = 1}^{K_1} {{u_k}{\theta _{ik}}{\theta _{jk}}}$ and then thresholding the drawn count at 1 through the indicator function ${\bf{1}}_{(\cdot)}$; 
the positive weight $u_k \in \mathbb{R}_+$ drawn from a gamma prior is used to balance the importance of the $k$th topic proportion pair $\theta_{ik} \theta_{jk}$ during the generation of the edge $a_{ij}$; the $k$th topic $\phiv_k \in \mathbb{R}_+^{K_1}$  drawn from a Dirichlet prior satisfies the simplex 
constraint, bringing the characteristics of scale identifiability and ease of inference \cite{zhou2016augmentable}.

Thanks to the conjugacy between the gamma and Poisson distributions, the developed GPFA in Eq.~\eqref{eq_GPFA} can estimate the model parameters with  analytic conditional posteriors, rather than adopting sophisticated approximate assumptions like existing RTMs \cite{chang2010hierarchical, liu2009topic}.
Moreover, the Bern-Poisson link \cite{DBLP:conf/aistats/Zhou15} in GPFA, which is used to model the generative process of edges $\Amat$, can be flexibly replaced with the Poisson likelihood or Gamma-Poisson link to model the generation of count or real positive edges, respectively, as discussed in Appendix A.

\subsection{Graph Poisson Gamma Belief Network}

To enhance the modeling capability and interpretability of existing RTMs \cite{chang2010hierarchical, liu2009topic}, which are often constrained by their employing single-layer network structures, we propose to construct a deep RTM with multiple stochastic hidden layers. 
To achieve the goal of exploring the multilevel semantics of document networks, one of the most straightforward ways to extend the GPFA in Eq.~\eqref{eq_GPFA} is to apply a hierarchical prior on its topic proportion $\thetav_j$ but fix the generative process of edges, following the regular procedure of extending a shallow topic model to a hierarchical one \cite{gan2015scalable, zhou2016augmentable}. 
However, the single-layer edge generative process ignores the relationships implied in multiple semantic levels, motivating us to construct a deep (hierarchical) RTM named Graph Poisson Gamma Belief Network (GPGBN), whose version with $T$ hidden layers can be formulated as
\begin{align} \label{eq_GPGBN}
&{\thetav} _j^{(T)} \sim \mbox{Gam}({\gammav} ,1/c_j^{(T + 1)}), \notag\\
&\cdots \notag\\
&{\thetav} _j^{(t)}\sim \mbox{Gam}({{\Phimat} ^{(t + 1)}}{\thetav} _j^{(t + 1)},1/c_j^{(t + 1)}), ~\phi_k^{(t)} \sim \mbox{Dir}(\eta^{(t)}),\notag\\
&{m_{ij}^{(t)}\sim \mbox{Pois}(\sum\nolimits_{k = 1}^{{K_t}} {u_k^{(t)}\thetav _{ik}^{(t)}\thetav _{jk}^{(t)}} )}, ~u_k^{(t)} \sim \mbox{Gam}({\alpha_k}, 1/\beta_k) \notag\\
&\cdots \notag\\
&{\xv}_j \sim \mbox{Pois}({{\Phimat} ^{(1)}}{\thetav} _j^{(1)}), \notag\\
&{a_{ij}}= 1({\delta _{ij}} \textgreater 1),\quad {\delta _{ij}} = \sum\nolimits_{t = 1}^T {m_{ij}^{(t)}}.
\end{align}
Similar to GPFA in Eq.~\eqref{eq_GPFA}, the observed document $\xv_j \in \mathbb{Z}_+^{K_0}$ is first factorized into the product of the topic matrix $\Phimat^{(1)} \in \mathbb{R}_+^{K_0 \times K_1}$ and the corresponding topic proportions (latent representation) $\thetav_j^{(1)} \in \mathbb{R}_+^{K_1}$ under the Poisson likelihood; then the $t$th-layer latent presentation $\thetav_j^{(t)} \in \mathbb{R}_+^{K_t}$ is factorized into the product of the topic matrix $\Phimat^{(t+1)} \in \mathbb{R}_+^{K_t \times K_{t+1}}$ and the topic proportions $\thetav_j^{(t+1)} \in \mathbb{R}_+^{K_{t+1}}$ at the next layer, denoting $K_t$ as the number of topics at the layer $t$; the binary edge $a_{ij}$ is generated by thresholding the summation of latent counts $\delta_{ij}=\sum\nolimits_{t = 1}^T {m_{ij}^{(t)}}$ at 1, where the $t$th-layer latent count $m_{ij}^{(t)}$ is drawn from the Poisson likelihood with a rate of $\sum\nolimits_{k = 1}^{{K_t}} {u_k^{(t)}\thetav _{ik}^{(t)}\thetav _{jk}^{(t)}}$; For ease of inference, we apply a Dirichlet prior on the topic $\phiv^{(t)} \in \mathbb{R}_+^{K_t}$ and a gamma prior on the importance weight $u_k^{(t)}$ at the layer $t$.

As shown in Fig.~\ref{fig_gpgbn} that provides an overview of the generative process, we can find that the developed GPGBN tightly couples the generative process of the node features $\Xmat$ and the adjacency matrix $\Amat$ by sharing their multilayer latent representations (unnormalized topic proportions) $ \{ \thetav_j^{(t)}\}_{j=1,t=1}^{N,T}$, making it possible to learn hierarchical semantic topics from document contents and also multilevel underlying relationships for each link (edge) between documents.
With the characteristics of Bern-Poisson link \cite{DBLP:conf/aistats/Zhou15}, the generative process of adjacency matrix can be equivalently reformulated by integrating $m_{ij}^{(t)}$ out as 
\begin{equation} \label{likelihood_a_PGBN}
a_{ij} \sim \mbox{Bernoulli} \left(1-\exp\left(-\sum_{t=1}^{T} \sum_{k = 1}^{{K_t}} {u_k^{(t)}\theta _{ik}^{(t)}\theta _{jk}^{(t)}}\right)\right). 
\end{equation}
Further, with the law of total expectation, we can estimate the expectation of $\xv_j$ and $a_{ij}$ under the Poisson likelihood as 
\begin{align}\label{eq:expectation}
\mathbb{E}\left[ \xv_j |- \right] &= \left[ \prod_{l=1}^t \Phimat^{(l)} \right] \ns \frac{\thetav_j^{(t)}}{\prod_{l=2}^t c_j^{(l)}},
\\\label{eq_expectation_a}
\mathbb{E}\left[ a_{ij} |- \right]&= 1-\exp\left(-\sum\nolimits_{t=1}^{T} \sum\nolimits_{k = 1}^{{K_t}} {u_k^{(t)}\theta _{ik}^{(t)}\theta _{jk}^{(t)}}\right), 
\end{align}
which reveals a series of appealing model properties as described in the following Section~\ref{subsec_model_properties}.

\subsection{Model  Properties} \label{subsec_model_properties}

\textbf{Hierarchical Semantic Topics:} 
The developed GPGBN in Eq.~\eqref{eq_GPGBN} inherits the characteristic of topic hierarchy from traditional deep topic models. 
Intuitively, from the view of expectation, Eq.~\eqref{eq:expectation} implies that the conditional expectation of $\xv_j$ is equivalent to a linear combination of the topics $\{{\prod_{l=1}^{t-1} \Phimat^{(l)} \phiv_k^{(t)}}\}_{k=1}^{K_t}$ with the corresponding document-dependent weights  $\{\theta_{jk}^{(t)}\}_{k=1}^{K_t}$ at the layer $t$. 
Therefore, the semantic meaning of $\phiv_k^{(t)}$ can be naturally interpreted with its projection to the observation space, denoted as ${\prod_{l=1}^{t-1} \Phimat^{(l)} \phiv_k^{(t)}}$, which provides us a principled way to visualize these learned topics at multiple semantic levels. 
As the hierarchical topic trees shown in Fig.~\ref{fig_topic_tree}, it is not difficult to find that the semantic meanings of topics learned by GPGBN tend to be specific at the bottom (lower) layer, and become more and more general when moving upwards (higher).

\textbf{Multilevel Semantic Relationships:} For the GPGBN formulated in Eq.~\eqref{eq_GPGBN}, 
after defining $a_{ij}^{(t)} := \sum\nolimits_{k = 1}^{{K_t}} {u_k^{(t)}\theta _{ik}^{(t)}\theta _{jk}^{(t)}}$, the encapsulated matrix $\Amat^{(t)} = \{a_{ij}^{(t)}\}_{i=1,j=1}^{N,N}$ can be treated as the ``\emph{adjancency matrix}'' at the layer $t$, which describes the specific semantic relationships constructed by the $t$th-layer latent document representations as visualized in Fig.~\ref{fig:relationships}. 
Further, by substituting $\{\Amat^{(t)}\}_{t=1}^{T}$ into Eq.~\eqref{eq_expectation_a}, we can have
\begin{align}
\mathbb{E}\left[ \Amat|- \right] = 1- \exp \left(- \sum_{t=1}^T \Amat^{(t)}\right),
\end{align}
which indicates that the observed adjacency $\Amat$ in GPGBN is generated by aggregating the semantic relationships across all hidden layers.
Moreover, from the expectation of $a_{ij}^{(t)}$ expressed in Eq.~\eqref{eq_expectation_a}, we can find that the positive scale parameter $u_k^{(t)}$ varies with the topic index $k$ and the layer index $t$, which implies that the contributions of various topics $\{\phiv_k^{(t)}\}_{k=1,t=1}^{K_t,T}$ to the generation of each edge are different.

\begin{figure}[!t]
	\centering
	 {\includegraphics[height=78mm]{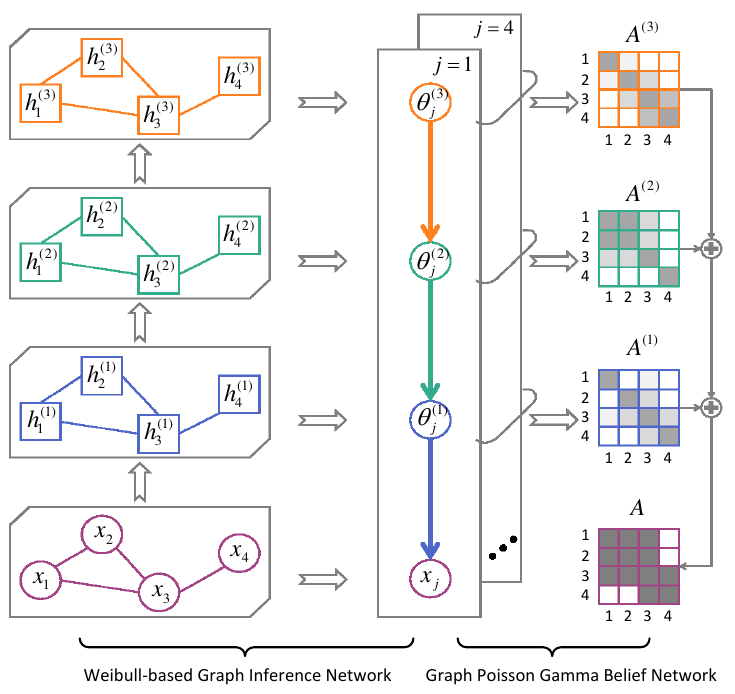}
		\label{generate}} 
	\caption{Illustration of the network structure of a 3-layer Weibull graph autoencoder (WGAE), which consists of a Weibull-based graph inference network (encoder) on the left and a GPGBN (decoder) on the right.}
	\label{model}
\end{figure}

\textbf{Effectiveness of Bern-Poisson link:} As shown in Eq.~\eqref{likelihood_a_PGBN}, as opposed to other link functions for binary observations, such as logistic or probit function, an appealing property of the Bern-Poisson link \cite{DBLP:conf/aistats/Zhou15} is that its inference cost only depends on the number of non-zero elements in the observed binary matrices \cite{zhou2015infinite}, making it efficient to deal with  large-scale sparse graphs.
Moreover, for each edge $a_{ij}$, the expectation formulation of Bern-Poisson link in Eq.~\eqref{likelihood_a_PGBN}, which is obtained by integrating the latent counts $\{m_{ij}^{(t)}\}_{t=1}^{T}$ out, provides us a straightforward way to directly calculate the gradients for the latent representations of the corresponding document pair, denoted as $\{\thetav_i^{(t)}$,$\thetav_j^{(t)}\}_{t=1}^T$, according to the loss function. 
These properties mentioned above make the inference of our models (including the GAE-liked extensions to be introduced below) be more efficient.

\textbf{Analytic Posteriors for Model Parameters:} Moving beyond existing RTMs' performing approximate parameter inference due to the non-conjugacy in generative models \cite{chang2010hierarchical, liu2009topic}, the developed GPGBN is constructed on the Poisson-gamma distribution and can provide analytic conditional posteriors for all model parameters, which can be directly inferred via Gibbs sampling.
One of the most important sampling update equtions, specifically the posterior of the topic proportion $\thetav_j^{(t)}$, can be formulated as
\begin{align} \label{eq_theta_posterior}
    p({\theta^{(t)}_{jk}} \given - )
    \sim&~\mbox{Gam}\big({x^{(t)}_{ \cdot jk}} + {\bm{\phi} ^{(t+1)}_{k:}}{\thetav} _j^{(t+1)} \ns+\ns \sum\limits_{i \ne j} {{m^{(t)}_{ijk}}}, \notag \\
    &{[-\ln(1-p^{(t)}_j) + {c^{(t+1)}_j} + {u^{(t)}_k}\sum\limits_{i \ne j} {{\theta^{(t)}_{ik}}} ]^{ - 1}} \big),
\end{align}

where ${x_{ \cdot jk}^{(t)}}$ and ${{m_{ijk}^{(t)}}}$ are the $t$th-layer latent count variables augmented from the corresponding node feature $\xv_j$ and relative edge $a_{ij}$, respectively. See Appendix B for a 
 detailed derivation.

\section{Weibull Graph Attention Autoencoder} \label{sec_wgcae}
Although the developed GPGBN equipped with analytic conditional posteriors can be efficiently trained with the Gibbs sampler accelerated with GPU \cite{CPGBN_ICML2019}, there are still several limitations, including:
$i$) restricted by training GPGBN with Gibbs sampling, it is not easy to plug in extra side information, such as the node labels;
$ii$) during the testing procedure, the developed GPGBN still relies on an iterative sampling process to extract the latent representations from documents, which is quite time-consuming;
$iii$) when applied to the practical graph analytic tasks with different focuses, GPGBN has difficulties in balancing the importance of  modeling the node features $\Xmat$ and adjacency matrix $\Amat$.

To move beyond these limitations, we consider to extend the developed GPGBN in Eq.~\eqref{eq_GPGBN} under the VAE framework \cite{kingma2014stochastic}.
As shown in Fig.~\ref{model}, taking the GPGBN as the generative network (decoder), we construct a Weibull-based graph inference network (encoder) to approximate the posterior of latent document representations $\{ \thetav_j^{(t)}\}_{j=1,t=1}^{N,T}$ in Eq.~\eqref{eq_theta_posterior}, and combine it with the GPGBN, leading to a  Weibull graph autoencoder (WGAE) with multiple stochastic hidden layers.
Then, inspired by $\beta$-VAE \cite{higgins2017beta}, we modify the loss function of the developed WGAE by introducing a hyper-parameter $\beta$ into the evidence lowerbound (ELBO) of data log-likelihood  $\log p(\Xmat, \Amat)$, which can be expressed as 
\begin{align} \label{eq_ELBO}
L= &\sum_{j = 1}^N {\mathbb{E}}\left[ {\ln p({\bm{x}_j}|{\bm{\Phi} ^{(1)}},\bm{\theta} _j^{(1)})} \right] \notag \\
&+ \beta {\mathbb{E}} \left[ \ln p(\bm{A}|\{ \thetav_j^{(t)}\}_{j=1,t=1}^{N,T} ) \right] \notag \\
&- \sum_{j = 1}^N {\sum_{t = 1}^T {{\mathbb{E}}\left[ {\ln \frac{{q(\bm{\theta} _j^{(t)}| - )}}{{p(\bm{\theta} _j^{(t)}|{\bm{\Phi} ^{(t + 1)}},\bm{\theta} _j^{(t + 1)})}}} \right]} },
\end{align}
where the hyper-parameter $\beta$ controls the importance balance between the likelihoods of node features $\Xmat$ and adjacency matrix $\Amat$. 
In the following, we will present two different network designs of the Weibull-based graph inference network (encoder), leading to two variants of WGAE, and develop the corresponding model inference algorithm for training.

\subsection{Weibull Graph Convolutional Autoencoder} \label{subsec_wgcae}

In details, given the global parameters $\{\Phimat^{(t)}, \uv^{(t)}\}_{t=1}^T$ (with $\uv^{(t)} = [u_1^{(t)}, \cdots, u_{K_t}^{(t)}]$) in Eq.~\eqref{eq_GPGBN}, the task of the graph inference network is to infer the local parameters of GPGBN, specifically $\{\thetav_j^{(t)}\}_{t=1}^T$ and $\{m_{ij}^{(t)}\}_{t=1}^T$.
And the fully factorized variational distribution estimated by the inference network can be represented as $\prod_{i,j=1}^N \prod_{t=1}^T q(\thetav_j^{(t)}|-) q(m_{ij}^{(t)}|-)$, where the posterior of $m_{ij}^{(t)}$ follows a Poisson distribution that is a discrete distribution and hard to optimize, as the derivation details shown in Appendix B.
Fortunately, benefiting from adopting Bern-Poisson link \cite{DBLP:conf/aistats/Zhou15}, the latent count $m_{ij}^{(t)}$ can be integrated from the likelihood of the $a_{ij}$ as shown in Eq.~\eqref{likelihood_a_PGBN}, resulting in that the graph inference network can be designed to only estimate $\prod_{j=1}^N \prod_{t=1}^T q(\thetav_j^{(t)}|-)$.

Moving beyond most existing VGAEs relying on Gaussian latent variables, inspired by Zhang et al. \cite{zhang2018whai}, we construct a Weibull-based graph inference network to approximate the gamma distributed conditional posterior of latent document representation, specifically $p(\thetav_j^{(t)}|-)$ in Eq.~\eqref{eq_theta_posterior}, formulated as 
\begin{equation} \label{eq_Weibull}
q(\bm{\theta} _j^{(t)}| - ) = \mbox{Weibull}(\bm{k}_j^{(t)} + {\bm{\Phi} ^{(t + 1)}}\bm{\theta} _j^{(t + 1)},\bm{\lambda} _j^{(t)}),
\end{equation}
where, the parameters $\bm{k}_j^{(t)},\bm{\lambda} _j^{(t)} \in {\mathbb{R}^{{K_t}}}$ are deterministically mapped from
the observed node features $\Xmat$ and adjacency matrix $\Amat$;
the product of $\Phimat^{(t+1)}$ and $\thetav_j^{(t+1)}$ injects the prior information passing from the higher layer $t+1$ into the $t$th-layer latent document representation $\thetav_j^{(t)}$.
We emphasize that the Weibull distribution has been proven to be an accurate approximation for the gamma distributed latent document representation, which is well known sparse, non-negative and difficult to be fitted with the Gaussain reparameterization. 
Further, constructing the inference network with the Weibull distribution can bring an efficient reparameterized formulation for $\thetav_j^{(t)}$ in Eq.~\eqref{eq_Weibull}, expressed as 
\begin{equation} \label{eq_reparameterize}
\thetav_j^{(t)} = \lambdav_j^{(t)} {( - \ln (1 - \varepsilonv_j^{(t)} ))^{1/(\bm{k}_j^{(t)} + {\bm{\Phi} ^{(t + 1)}}\bm{\theta} _j^{(t + 1)})}},
\end{equation}
where $\varepsilonv_j^{(t)} \sim \mbox{Uniform}(0,1),$
leading to an expedient and numerically stable gradient calculation.

For the detailed inference network structure, we adopt the recent popular GNNs to realize these transformations, one based on the vanilla GCN \cite{kipf2016semi} as shown in below and the other based on the Bayesian attention mechanism \cite{fan2020Bayesian} as shown in Section~\ref{subsec_wgaae}. 

\textbf{Weibull-based Graph Convolutional Encoder:} 
Attracted by the excellent characteristics of GCN \cite{kipf2016semi} in aggregating and propagating graph structure information, in the first variant, 
we use the vanilla GCNs to construct the deterministic transformation from $\{\Xmat, \Amat\}$ to $\{\kv_j^{(t)}, \lambdav _j^{(t)} \}_{j=1,t=1}^{N,T}$ in Eq.~\eqref{eq_Weibull}, formulated as
\begin{align}
&\Hmat^{(t)}=\ln [1 + \exp (\tilde{\Amat}\Hmat^{(t - 1)}\Wmat_1^{(t)})], &\quad t=1, \cdots, T  \notag\\
&\Kmat^{(t)} = \ln [1 + \exp (\tilde{\Amat}\Hmat^{(t)}\Wmat_2^{(t)})], &\quad t=1, \cdots, T \\
&\Lambdamat^{(t)}= \ln [1 + \exp (\tilde{\Amat}\Hmat^{(t)}\Wmat_3^{(t)})], &\quad  t=1, \cdots, T \notag
\end{align}
where, $\Hmat^{(0)} = \Xmat^T \in \mathbb{R}^{N \times K_0}$ denotes the input node features; $\Kmat^{(t)} = [\kv_1^{(t)};\cdots;\kv_N^{(t)}] \in \mathbb{R}^{N \times K_t}$ and $\Lambdamat^{(t)} = [\lambdav_1^{(t)};\cdots;\lambdav_N^{(t)}] \in \mathbb{R}^{N \times K_t}$ are the shape and scale parameters of the Weibull distribution, respectively; $\{\Wmat_i^{(t)}\}_{i=1}^{3} \in \mathbb{R}^{K_{t-1} \times K_t}$ denote the convolutional filters of $t$th-layer GCNs; $\Hmat^{(t)} \in \mathbb{R}^{N \times K_t}$ represents the node embedding at the layer $t$;
$\tilde{\bm{A}}={\bm{Q}^{ - \frac{1}{2}}}\bm{A}{\bm{Q}^{ - \frac{1}{2}}}$ is the normalized symmetric adjacent matrix shared across all layers with the degree matrix $\bm{Q}$.

We refer to the combination of the probabilistic decoder GPGBN and the Weibull-based graph convolutional encoder as Weibull graph convolutional autoencoder (WGCAE). Distinct from existing VGAEs based on GCNs \cite{hasanzadeh2019semi, yin2018semi}, the developed WGCAE owns the following model characteristics: $i)$ take a probabilistic generative model as the decoder to model the generation of node features and adjacency matrix jointly, where existing VGAEs only focus on the generation of edges; 
$ii)$ use multiple stochastic hidden layers to extract the latent document representation on various semantic levels;
$iii)$ adopt a Weibull reparameterization rather than a Gaussian one to approximate the posterior of gamma distributed latent document representations.

\subsection{Weibull Graph Attention Autoencoder} \label{subsec_wgaae}

For the second variant, we try to construct a graph inference network based on the recently proposed Bayesian attention mechanism \cite{fan2020Bayesian}. 
Distinct from GCN's fixed normalized symmetric adjacent matrix $\tilde{\Amat} \in \mathbb{R}_+^{N \times N}$ for massage passing, where each element is completely determined by the observed edge $a_{ij}$ and the degrees of the corresponding connected nodes, we construct a novel Weibull-based graph attention encoder based on the Bayesian attention mechanism, which can provide both stochastic attention weights $\{\Smat^{(c, t)} \in \mathbb{R}_+^{N \times N}\}_{c=1, t=1}^{C, T}$ and latent document representations $\{\thetav_j^{(t)}\}_{t=1}^{T}$, denoting $C$ as the number of heads in the attention module.

{\bf{Weibull-based Graph Attention Encoder:}} Specifically,  for each head at the layer $t$, we can first obtain the unnormalized attention weights $\Mmat^{(c, t)} = \{m_{ij}^{(c, t)}\}_{i=1,j=1}^{N,N}$ by mapping from the output of the previous layer, specifically $\Hmat \in \mathbb{R}_+^{N \times K_t}$, where each coefficient $m_{ij}^{(c, t)}$ can be computed as 
\begin{align}
    m_{ij}^{(c, t)} =\text{LeakyRelu}({\av^{(t)}}^T([\Wmat^{(t)}\hv_i^{(t)}\Vert\Wmat^{(t)}\hv_j^{(t)}])),
\end{align}
where $\Wmat^{(t)} \in \mathbb{R}^{K_t \times K_t}$ and $\Vert$ is the concatenation operation; $\av \in \mathbb{R}^{2K_t}$ {stands for a weighted vector to project the concatenated representation to the corresponding attention~score.}

Following that, inspired by Fan et al. \cite{fan2020Bayesian}, we construct a stochastic Weibull-based variational attention layer to transform these deterministic attention weights $\Mmat^{(c, t)} = \{m_{ij}^{(c, t)}\}_{i=1,j=1}^{N,N}$ into a stochastic attention distribution, formulated as
\begin{align}
    &s_{ij}^{(c,t)} = \exp(m_{ij}^{(c,t)})\frac{-\log(1-\epsilon_{i,j}^{(c,t)})^{1/k}}{\Gamma(1+1/k)},
\end{align}
where $\epsilon_{i,j}^{(c,t)} \sim \text{Uniform}(0, 1)$, $k$ is a global hyperparameter, and $\Gamma (\cdot)$ denotes the gamma function.
These obtained stochastic attention weights $\Smat^{(c,t)}=\{s_{ij}^{(c,t)}\}_{i=1,j=1}^{N,N}$ can be treated equivalently as sampled from the distribution $ \text{Weibull}(k, \frac{\exp(\Mmat^{(c,t)})}{\Gamma(1+1/k)})$, and each element $s_{ij}^{(c,t)}$ can be further normalized as 
\begin{align}
    \hat s_{ij}^{(c,t)} = \frac{\exp(s_{ij}^{(c,t)})}{\sum_{j \ \in \mathcal{N}_i}\exp(s_{ij}^{(c,t)})},
\end{align}
where $\mathcal{N}_i$ denotes the set of nodes connected to the $i$th one. 
With the stochastic attention weights after normalization, specifically $\hat \Smat^{(c,t)}=\{\hat s_{ij}^{(c,t)}\}_{i=1,j=1}^{N,N}$, we build the transformation from $\{\Xmat, \Amat\}$ to $\{\kv_j^{(t)}, \lambdav _j^{(t)} \}_{j=1,t=1}^{N,T}$ in Eq.~\eqref{eq_Weibull}, formulated as
\begin{align}
&\Hmat^{(t)} = \frac{1}{C}\sum\limits_{c=1}^{C}\hat \Smat^{(c,t)}\Hmat^{(t-1)}\Wmat_1^{(c,t)}    &\quad t=1, \cdots, T \notag\\
&\Kmat^{(t)} = \ln [1 + \exp (\Hmat^{(t)} \Wmat_2^{(t)})], &\quad t=1, \cdots, T \\
&\Lambdamat^{(t)} = \ln [1 + \exp (\Hmat^{(t)} \Wmat_3^{(t)})], &\quad  t=1, \cdots, T \notag
\end{align}
where, $\Hmat^{(0)} = \Xmat^T \in \mathbb{R}^{N \times K_0}$ denotes the input node features; $\Kmat^{(t)} = [\kv_1^{(t)};\cdots;\kv_N^{(t)}] \in \mathbb{R}^{N \times K_t}$ and $\Lambdamat^{(t)} = [\lambdav_1^{(t)};\cdots;\lambdav_N^{(t)}] \in \mathbb{R}^{N \times K_t}$ are the shape and scale parameters of Weibull distribution respectively; 
$\{\Wmat_i^{(t)}\}_{i=1}^{3} \in \mathbb{R}^{K_{t-1} \times K_t}$  denote the projection matrix at the layer $t$; $\Hmat^{(t)} \in \mathbb{R}^{N \times K_t}$ represents the node embedding at the layer $t$.

We refer to the combination of the probabilistic decoder GPGBN and the Weibull-based graph attention encoder as Weibull graph attention autoencoder (WGAAE). 
Distinct from the deterministic attention mechanism in GAT \cite{velickovic2018graph}, the Bayesian attention mechanism in the Weibull-based graph attention encoder treats the attention weights as latent random variables, which can introduce the uncertainty estimation to bring more robustness \cite{fan2020Bayesian}. 
Moreover, compared to GCN's fixed normalized symmetric adjacent matrix $\tilde{\Amat} \in \mathbb{R}_+^{N \times N}$ for massage passing, the developed WGAAE can leverage multi-head trainable propagation weights and significantly improve the model performance with the self-attention mechanism.








\subsection{Full-batch Training} \label{subsec_end2end}

Combining the probabilistic decoder GPGBN in Eq.~\eqref{eq_GPGBN} with the previous Weibull-based graph inference networks, we have developed two variants of WGAE, including WGACE in Section~\ref{subsec_wgcae} and WGAAE in Section~\ref{subsec_wgaae}.
In what follows, a hybrid Bayesian inference algothrim is employed to learn all parameters $\Omegamat = \{ \uv = \{u_k^{(t)}\}_{k=1,t=1}^{K_t,T}, \{\Phimat^{(t)}\}_{t=1}^T, \Wmat_e\}$ in our models, where $\{u_k^{(t)}\}_{k=1,t=1}^{K_t,T}$ and $\{\Phimat^{(t)}\}_{t=1}^T$ are global parameters defined in GPGBN (decoder) and $\Wmat_e$ encapsulates the parameters in the graph inference network (encoder).

Aimed at training both the encoder and decoder parts of WGAEs jointly, we first develop a full-batch training algorithm, which can alternately update $\{\Phimat^{(t)}\}_{t=1}^T$ with Gibbs sampling and $\{\{u_k^{(t)}\}_{k=1,t=1}^{K_t,T}$, $\Wmat_e\}$ with GD-based optimization method, resulting in an end-to-end training procedure.
As shown in Algorithm~\ref{algorithm_1}, for each iteration, the developed hybrid Gibbs sampling and variational inference will first draw random noise $\bm{\varepsilon}  = \{ \bm{\varepsilon} _j^{(t)}\} _{j = 1,t = 1}^{N_s,T}$ for all documents, and then update $\{\{u_k^{(t)}\}_{k=1,t=1}^{K_t,T}$, $\Wmat_e\}$ with the gradients ${\nabla _{{\Wmat_e},\bm{u}}}L({\Wmat_e}, \bm{u}; \Xmat, \Amat,\bm{\varepsilon})$ to maximize the loss function defined in Eq.~\eqref{eq_ELBO}. 
After sampling the Weibull-distributed latent document representations $\{ \bm{\theta} _j^{(t)}\} _{j = 1,t = 1}^{N,T}$ with Eq.~\eqref{eq_Weibull}, the updates for $\{\Phimat^{(t)}\}_{t=1}^T$ equipped with analytic posteriors can be processed with Gibbs sampling, which can be further accelerated with GPU.
The proposed hybrid Gibbs sampling and variational inference algorithm is implemented in TensorFlow \cite{abadi2016tensorflow} combined with pyCUDA \cite{klockner2012pycuda} for efficient computation.

\begin{algorithm}[t] 
\caption{Hybrid Gibbs sampling and variational inference for WGCAE and WGAAE}
\begin{algorithmic}
\STATE{Set the number of hidden layers ${T}$;}
\STATE{Initialize parameters $\{ \{u_k^{(t)}\}_{k=1,t=1}^{K_t,T}, \{\Phimat^{(t)}\}_{t=1}^T, \Wmat_e\}$;}  
\FOR{$iter = 1,2,\cdots$}
\STATE{For all documents $\Xmat \in \mathbb{Z}^{K_0 \times N}$ and $\Amat \in {\{ 0,1\} ^{N \times N}}$
draw random noise $\bm{\varepsilon}  = \{ \bm{\varepsilon} _j^{(t)}\} _{j = 1,t = 1}^{N,T}$;}
\STATE{Calculate ${\nabla _{{\Wmat_e},\bm{u}}}L({\Wmat_e}, \bm{u}; \Xmat, \Amat,\bm{\varepsilon})$ according to Eq.~\eqref{eq_ELBO} and update $\{\bm{u} = \{u_k^{(t)}\}_{k=1,t=1}^{K_t,T}, \Wmat_e\}$;}
\STATE{Sample corresponding topic proportions $\{ \bm{\theta} _j^{(t)}\} _{j = 1,t = 1}^{N,T}$ to update $\{\Phimat^{(t)}\}_{t=1}^T$ with Gibbs sampling;
}
\ENDFOR
\end{algorithmic}
\label{algorithm_1}
\end{algorithm}

\begin{algorithm}[t] 
\caption{Hybrid stochastic-gradient MCMC and variational inference for WGCAE and WGAAE}
\begin{algorithmic}
\STATE{Set the mini-batch size $N_s \le N$ and the number of hidden layers ${T}$;}
\STATE{Initialize parameters $\{ \{u_k^{(t)}\}_{k=1,t=1}^{K_t,T}, \{\Phimat^{(t)}\}_{t=1}^T, \Wmat_e\}$;}
\FOR{$iter = 1,2,\cdots$}
\STATE{Select a mini-batch of $N_s$ documents with the probabilities defined in Eq.~\eqref{eq_node_sampling} to form a subset
$\Xmat_s = \{ {\bm{x}_j}\} _{j = 1}^{N_s}$;}
\STATE{For the selected documents $\Xmat_s \in \mathbb{Z}^{K_0 \times N_s}$ and $\Amat_s \in {\{ 0,1\} ^{N_s \times N_s}}$
draw random noise $\bm{\varepsilon}  = \{ \bm{\varepsilon} _j^{(t)}\} _{j = 1,t = 1}^{N_s,T}$;}
\STATE{Calculate ${\nabla _{{\Wmat_e},\bm{u}}}L({\Wmat_e}, \bm{u}; \Xmat_s, \Amat_s,\bm{\varepsilon})$ according to Eq.~\eqref{eq_ELBO} and update $\{\bm{u} = \{u_k^{(t)}\}_{k=1,t=1}^{K_t,T}, \Wmat_e\}$;}
\STATE{Sample corresponding topic proportions $\{ \bm{\theta} _j^{(t)}\} _{j = 1,t = 1}^{N_s,T}$ to update $\{\Phimat^{(t)}\}_{t=1}^T$ with SG-MCMC in Eq.~\eqref{eq_online};
}
\ENDFOR
\end{algorithmic}
\label{algorithm_2}
\end{algorithm}

\subsection{Scalable Training} \label{scalable}

Considering the limited scalability of processing all documents in each iteration and the ubiquitous of large graphs with millions of nodes,  we propose a scalable training algorithm for WGAEs as shown in Algorithm~\ref{algorithm_2}, which updates the model parameters with mini-batch estimation rather than full-batch estimation.
In detail, during the full-batch training of GAEs, the computation cost of the likelihood of the entire adjacency matrix $\Amat$ suffers from a quadratic complexity and is unaffordable when dealing with large-scale graphs.
To this end, we follow Salha et al. \cite{DBLP:journals/nn/SalhaHRMV21} to subsample $N_s \le N$ nodes from the original graph and estimate the gradients of model parameters with a smaller graph $\Amat_s \in \{0,1\}^{N_s \times N_s}$, which greatly reduces the computation cost in each iteration and makes the developed WGAE scalable.

Distinct from the subsampling method in FastGAE \cite{DBLP:journals/nn/SalhaHRMV21}, we develop a more flexible strategy to sample nodes from the original graph with replacement. Specifically, we sample the node $i$ with the acceptance probability $p_i$, formulated as:
\begin{align} \label{eq_node_sampling}
    p_i = k q_i  + (1-k) \frac{1-q_i}{N-1}, ~q_i = \frac{f(\xv_i)^\alpha}{\sum_{j=1}^N(f(\xv_j)^\alpha)},
\end{align}
where $f(\cdot)$ measures the ``\emph{importance}'' of node $i$ and $q_i$ denotes the probability of sampling node $i$ for its relative importance in the original graph. 
Likewise, $\frac{1-q_i}{N-1}$ can be interpreted as the probability that the node $i$ is sampled for its relative unimportance. For a better coverage, both ``\emph{important nodes}'' and ``\emph{unimportant nodes}'' are to be included in the sampled subgraph, with their percentages balanced by $k\in[0,1]$, \emph{i.e.}, the higher the value of $k$, the more ``\emph{important nodes}'' are expected to be included in the subgraph. In practice, we adopt the node degree as $f(\cdot)$ in our experiments and the sharpness of the probability distribution $\{q_i\}_{i=1,N}$ can be adjusted by the exponent $\alpha$, where greater sharpness can be obtained with a higher value of $\alpha$, and $\{q_i\}_{i=1,N}$ would be equivalent to an uniform distribution if $\alpha$ is set to 0.


Based on the node sampling algorithm in Eq.~\eqref{eq_node_sampling}, the model parameters of WGAEs, specifically $\Omegamat = \{ \uv = \{u_k^{(t)}\}_{k=1,t=1}^{K_t,T}, \{\Phimat^{(t)}\}_{t=1}^T, \Wmat_e\}$, can be updated with a hybrid stochastic-gradient MCMC and variational inference as shown in Algorithm~\ref{algorithm_2}.
For each iteration, $\{\{u_k^{(t)}\}_{k=1,t=1}^{K_t,T}$, $\Wmat_e\}$ can be updated with the SGD-based optimization method, which calculates the stochastic gradients ${\nabla _{{\Wmat_e},\bm{u}}}L({\Wmat_e}, \bm{u}; \Xmat_s, \Amat_s,\bm{\varepsilon})$ based on the mini-batch consisting of selected documents $\Xmat_s = \{ {\bm{x}_j}\} _{j = 1}^{N_s}$.
The Gibbs sampling for $\{\Phimat^{(t)}\}_{t=1}^T$ can be replaced with the extension of TLASGR-MCMC \cite{cong2017deep}, which is a SG-MCMC algorithm and has been flexibly applied for the scalable inference for discrete latent variable models (LVMs). In detail, the efficient TLASGR-MCMC update for each ${\bm{\phi}}_k^{(t)} \in \mathbb{R}_+^{K_t}$ in WGAEs can be formulated as
\begin{align}  \label{eq_online}
&{\bm{\phi}}_k^{(t)new} = \Big\{{\bm{\phi}}_k^{(t)} + \frac{{\varepsilon _i^{(t)}}}{{\bm{M}_k^{(t)}}}[(\rho \bm{\widetilde x}_{:k \cdot }^{(t)} + {\bm{\eta} ^{(t)}}) \\
&{-(\rho \bm{\widetilde x}_{ \cdot k \cdot }^{(t)} + {\bm{\eta} ^{(t)}} {{K_{t - 1}}}){\bm{\phi} _k^{(t)}}] + N\big(0,\frac{{2{\varepsilon _i^{(t)}}}}{\bm{M}_k^{(t)}} \mbox{diag}({\bm{\phi}}_k^{(t)})\big)\Big\}_\angle}, \notag
\end{align}
where the subscript $i$ indicates the number of mini-batches processed so far; 
$\bm{\widetilde x}_{:k \cdot }^{(t)} \in \mathbb{Z}^{K_t}$ denotes the latent count vector augmented from the mini-batch $\Xmat_s$ with the symbol ``$\cdot$'' in the subscript indicating the summation across the sample dimension; and the definitions of $\rho$, $\varepsilon _i^{(t)}$, ${\bm{M}_k^{(t)}}$ and $\{\bm \cdot\}_\angle$ are analogous to these in Cong et al. \cite{cong2017deep} and omitted here for brevity.
For efficient computation, the proposed hybrid stochastic-gradient MCMC/autoencoding variational inference is also implemented in TensorFlow \cite{abadi2016tensorflow} combined with pyCUDA \cite{klockner2012pycuda}.

\section{Experiments} \label{sec_experiments}

\bb{To evaluate the effectiveness of the developed GPGBN and its VAE-liked extensions, specifically WGCAE and WGAAE, in what follows, we design a series of experiments including both qualitative analysis and quantitative comparisons. The source code has been released at \url{https://github.com/chaojiewang94/WGAAE}} and also been included in \url{https://github.com/BoChenGroup/PyDPM.}

\subsection{Qualitative Analysis}

\bb{At first, we focus on evaluating the model properties of GPGBN as discussed in Section~\ref{subsec_model_properties}, and design a series of qualitative experiments to intuitively investigate the hierarchical semantic topics and multilevel document relationships captured by GPGBN.
We emphasize that the further developed WGAEs also inherit these attractive model properties from GPGBN, benefiting from taking it as the generative network (decoder).}

\begin{figure*}[t]
\centering
\subfigure[]{\includegraphics[width=160mm]{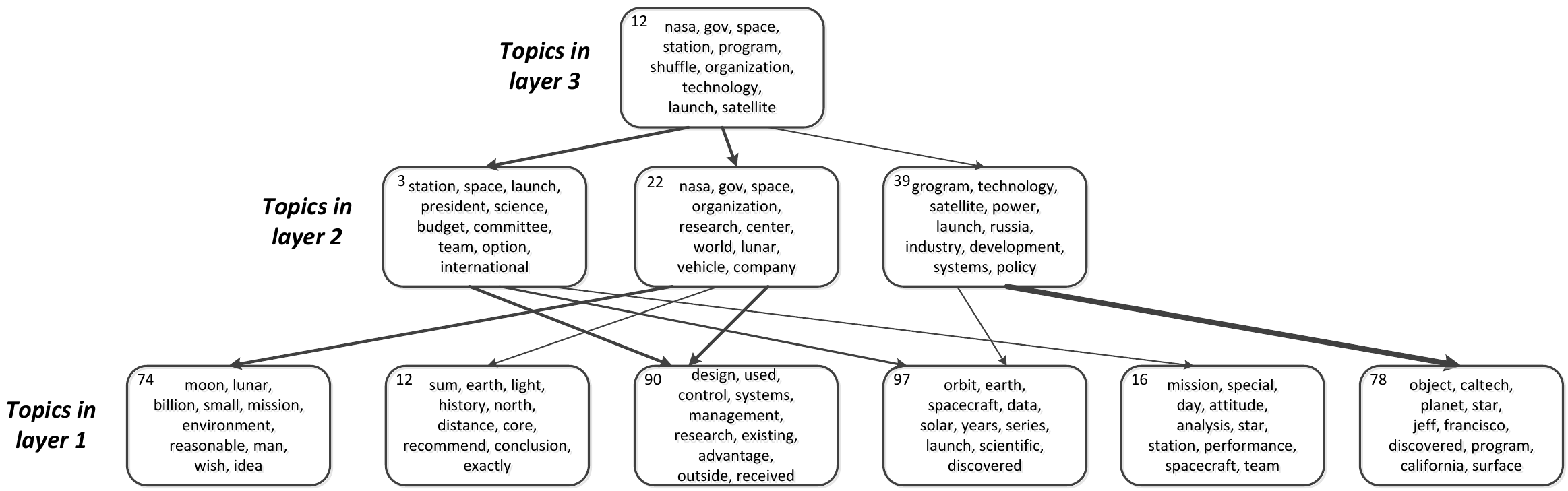}}
\subfigure[]{\includegraphics[width=160mm]{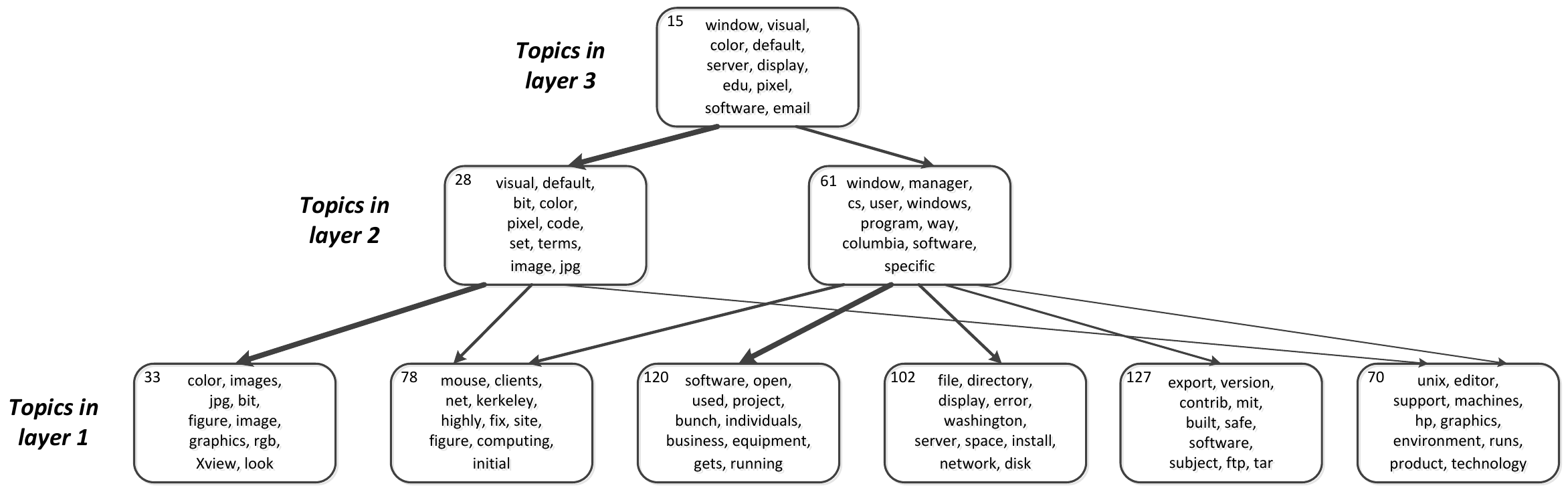}}
\caption{Visualization of hierarchical semantic topics learned by 3-layer GPGBNs on 20News dataset.}
\label{fig_topic_tree}
\end{figure*}

\textbf{Datasets \& Preprocess}

\bb{For qualitative analysis, we choose 20News \cite{20news} dataset as the benchmark, which consists of 18,774 documents collected from 20 different news groups and has been widely used for visualizing hierarchical network structures learned by deep topic models. 
Following the experimental settings in Zhou et al. \cite{zhou2016augmentable}, we represent each document as a BoW vector, denoted as $\xv_i \in \mathbb{Z}^{K_0}$ with a pre-pruned vocabulary of size $K_0=2000$.
Considering that there is not an existing adjacency matrix for 20News, we construct a hand-crafted binary one $\Amat= \{0,1\}_{i=1,j=1}^{N,N}$ by measuring the cosine similarity between documents and then threshold each document relationship $a_{ij}$ with 
\begin{equation}\label{Eq: adjacent matrix}
a_{ij}=\left\{
\begin{array}{ll}
1  & \text{if $\mbox{cos}(\bm{x}_i,\bm{x}_j) \ge \tau_a$} \\
0  & \text{if $\mbox{cos}(\bm{x}_i,\bm{x}_j) < \tau_a$},
\end{array} \right.
\end{equation}
where $\tau_a$ is a hyperparameter to adjust the sparsity of the adjacency matrix.}

\begin{table}[t]
    \centering
    \caption{Hyperparameters settings of GPGBN and WGAEs.}
    \label{tab:model_setting}
    \scalebox{0.85}{
    \begin{tabular}{c|c|c}
    \toprule
    Parameter Priors & Parameter Meanings& Hyperparameter Settings \\
    \midrule
    $\bm{\phi} _{{k_t}}^{(t)}\sim \mbox{Dir}({\eta ^{(t)}})$ &The $k_t$ topic at layer $t$  &$\eta ^{(t)}=0.01$ \\
    $c_j^{(t)}\sim \mbox{Gam}({e_0},1/{f_0})$ &The scale parameter of $\bm{\theta}_j$ & $e_0=1, f_0=1$ \\
    $u_{{k_t}}^{(t)}\sim \mbox{Gam}({\alpha _0},1/{\beta _0})$ &The importance weight of $\theta _{{k_t}j}^{(t)}$ & $\alpha_0=1$, $\beta _0=1$\\
    \bottomrule
    \end{tabular}}
\end{table}

\textbf{Model Settings} 

For network structures, we construct 3-layer GPGBNs with the same network structure of $[{K_1},{K_2},{K_3}] = [128,64, 32]$ from shallow to deep, and then train these GPGBNs with Gibbs sampling after 1000 iterations on each news group of 20News dataset.

For hyper-parameters, we list a tabular overview of parameter names, priors, meanings, and corresponding settings in Table~\ref{tab:model_setting} to make an intuitive introduction.
Here we note that the hyperparameter settings in Table~\ref{tab:model_setting} are suitable for all our models in the following experiments, including GPFA/GPGBN and WGCAE/WGAAE.

\textbf{Hierarchical Topic Visualization}

Following the procedure in Zhou et al. \cite{zhou2016augmentable}, we first visualize the hierarchical semantic topics learned by GPGBN on 20News dataset. 
For each topic tree shown in Fig.~\ref{fig_topic_tree}, we pick a node at the top layer as the root node and then grow the topic tree downward by drawing a line from the $k$th node at layer $t$ to the $k'$th node at layer $t-1$, for all node pairs satisfying the constraint $\{ k':\bm{\Phi} _{k'k}^{(t)} > {\tau_{\phi}^{(t)}}{\rm{/}}{K_{t - 1}}\}$, where ${\tau_{\phi}^{(t)}}$ is the hyper-parameter to adjust the complexity of the topic tree. 
For each exhibited topic node, we list the top-10 words according to their probabilities assigned to this topic to intuitively interpret its semantics.
From the results of visualization, we can find that the developed GPGBN can capture hierarchical topics at multiple semantic levels, where the topic semantics vary from specific to general with the increase of the network depth.

\begin{figure*}[t]
\centering
\subfigure[]{\includegraphics[width=160mm]{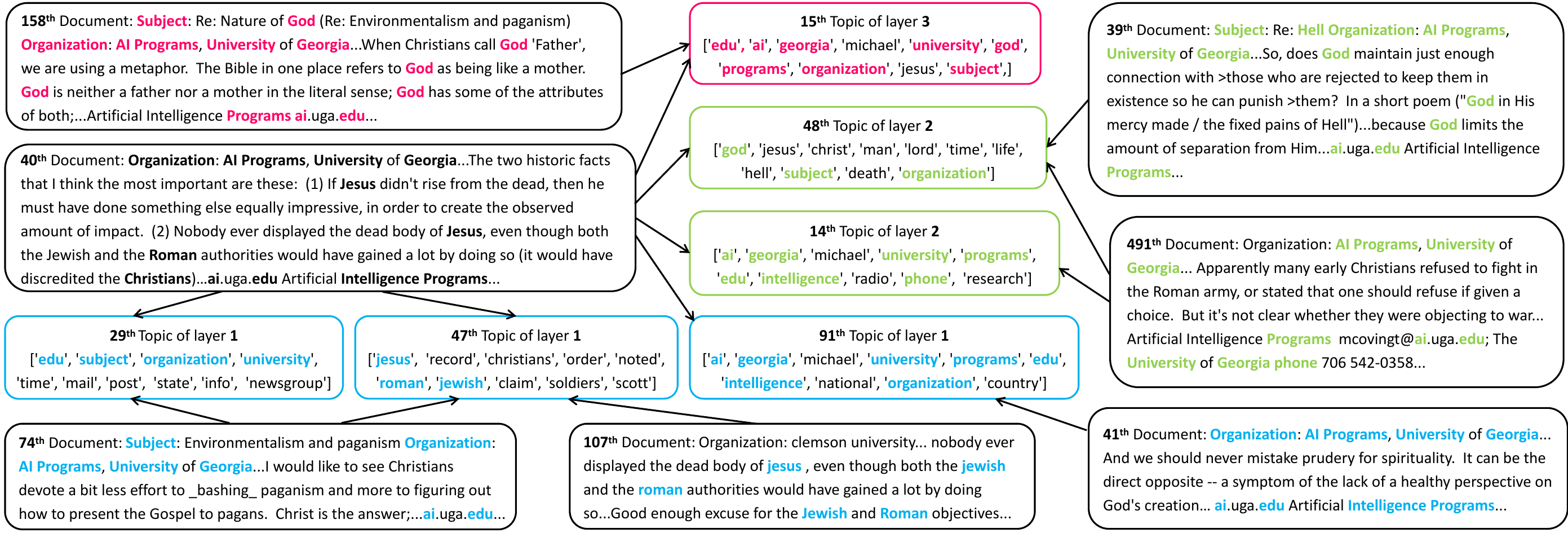} \label{figure:relation}} 
\subfigure[]{\includegraphics[width=160mm]{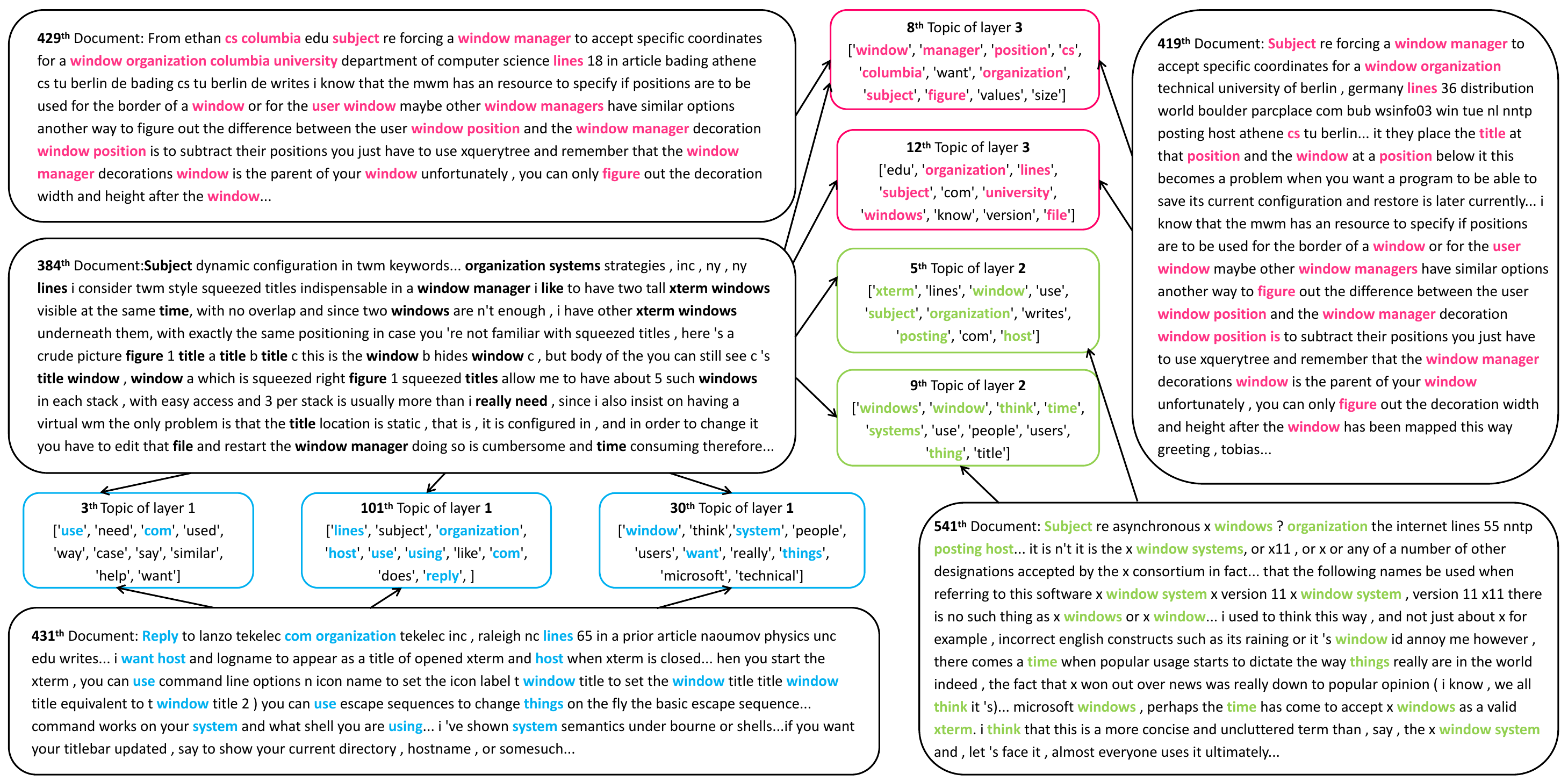} \label{figure:relation1}} 
\caption{Visualization of document subnetworks learned by a 3-layer GPGBN on 20News dataset.
For each subnetwork, taking the $i$th document as the source node, other documents, whose connections at layer $t$ (denoted as ${u_k^{(t)}\theta _{ik}^{(t)}\theta _{jk}^{(t)}}$) are larger than a threshold $\tau_u$, are displayed in the black boxes, and the key words of connected topics are displayed in the blue, green, and red text boxes, respectively, from shallow to deep. The common key words simultaneously appear in both documents connected with their associated topics are highlighted with the corresponding topic color.}
\end{figure*}

\textbf{Multilevel Relationship Visualization}

Distinct from all existing RTMs, the developed GPGBN can provide an explicit solution to explore multilevel semantic relationships between documents as discussed in Section~\ref{subsec_model_properties}. 
Taking the subnetwork exhibited in Fig.~\ref{figure:relation} as an example, after selecting the $40$th document as the source node, we exhibit other related documents, whose semantic connections to the $40$th document satisfy the constraint ${u_k^{(t)}\theta _{ik}^{(t)}\theta _{jk}^{(t)} > \tau_u }$, where ${u_k^{(t)}\theta _{ik}^{(t)}\theta _{jk}^{(t)}}$ measures the semantic relevance between the $i$th and $j$th documents at the $k$th topic of layer $t$ and $\tau_u$ is the hyper-parameter to adjust the complexity of the corresponding subnetwork.
For intuitive visualization, we display the content of each document in the corresponding black text box and also visualize the selected topics at multiple semantic levels with their key words.
We highlight the key words simultaneously appearing in both document content and connected topics with different colors, specifically blue, green, and red from shallow to deep, which potentially illustrates the reason why the semantics of two documents are connected through the corresponding topics.
For instance, the $40$th document is related with the $41$th, $107$th, and $74$th documents on specific topics at the first layer, like $29$th topic on ``\emph{edu}, \emph{organization}, \emph{university}'' and $47$th topic on ``\emph{jesus}, \emph{christians}, \emph{jewish},'' and has connections with other documents at higher semantic levels, where the semantics of topics become more and more general with the network depth going deeper.
Similar conclusions can be achieved by examining the other subnetworks shown in Fig.~\ref{figure:relation1}.

\subsection{Quantitative comparison}

\begin{table}[t]
    \centering
    \caption{Statistics of graph benchmarks for various tasks.}
    \label{tab:datasets}
    \scalebox{0.9}{
    \begin{tabular}{c|c|cccc}
    \toprule
    Task & Dataset &Nodes &Edges &Features &Classes\\
    \midrule
    link prediction & Cora  &2,708 &5,429  &1,433  &7\\
    or &Citeseer &3,327 &4,732 &3,703  &6\\
    node classification & Pubmed &19,717 &44,338 &500   &3\\
    \midrule
    & Coil &1,440  &4,201  &1,024 &20\\
    node clustering & TREC &5952 &18,013 &2000  &8 \\
    & R8 &7674 &27,513 &2000  &8 \\
    \bottomrule
    \end{tabular}}
\end{table}

To make a quantitative comparison with other popular RTMs and GAEs, in what follows, we evaluate the model performance of the developed WGAEs on a series of graph analytic tasks, including link prediction, node clustering, and classification.

\begin{table*}[t]
    \centering
    \caption{Comparisons of link prediction performance.}
    \label{tab:link_predict}
    \scalebox{1}{
    \begin{tabular}{c|cc|cc|cc}
    \toprule
    \textbf{Method} &\multicolumn{2}{|c}{\textbf{Cora}} &\multicolumn{2}{|c}{\textbf{Citeseer }} &\multicolumn{2}{|c}{\textbf{Pubmed}} \\
    &AUC &AP &AUC &AP &AUC &AP \\
    \midrule
    SC \cite{tang2011leveraging} &84.6$\pm$0.01 &88.5$\pm$0.00 &80.5$\pm$0.01 &85.0$\pm$0.01 &84.2$\pm$0.02 &87.8$\pm$0.01 \\
    DW \cite{perozzi2014deepwalk} &83.1$\pm$0.01 &85.0$\pm$0.00 &80.5$\pm$0.02 &83.6$\pm$0.01 &84.4$\pm$0.00 &84.1$\pm$0.00 \\
    GAE \cite{kipf2016variational} &91.0$\pm$0.02 &92.0$\pm$0.03 &89.5$\pm$0.04 &89.9$\pm$0.05 &96.4$\pm$0.00 &96.5$\pm$0.00 \\
    VGAE \cite{kipf2016variational} &91.4$\pm$0.01 &92.6$\pm$0.01 &90.8$\pm$0.02 &92.0$\pm$0.02 &94.4$\pm$0.02 &94.7$\pm$0.02 \\
    SEAL \cite{zhang2018link} &90.1$\pm$0.1 &83.0$\pm$0.3 &83.6$\pm$0.2 &77.6$\pm$0.2 &96.7$\pm$0.1 &90.1$\pm$0.1 \\
    G2G \cite{bojchevski2017deep} &92.1$\pm$0.9 &92.6$\pm$0.8 &95.3$\pm$0.7 &95.6$\pm$0.7 &94.3$\pm$0.3 &93.4$\pm$0.5 \\

    \midrule
    $S$-VGAE \cite{davidson2018hyperspherical} &94.1$\pm$0.1 &94.1$\pm$0.3 &94.7$\pm$0.2 &95.2$\pm$0.2 &96.0$\pm$0.1 &96.0$\pm$0.1 \\
    NF-VGAE \cite{hasanzadeh2019semi}&92.4 $\pm$0.6 &93.0$\pm$0.5 &91.8$\pm$0.3 &93.0$\pm$0.8 &96.6$\pm$0.3 &96.7$\pm$0.4\\
    Naive SIG-VAE  \cite{hasanzadeh2019semi}&94.0$\pm$0.5 &93.3$\pm$0.4 &94.3$\pm$0.8 &93.6$\pm$0.9 &96.5$\pm$0.7 &96.0$\pm$0.5 \\
    SIG-VAE (IP)  \cite{hasanzadeh2019semi}&94.4$\pm$0.1 &94.4$\pm$0.1 &95.9$\pm$0.1 &95.4$\pm$0.1 &96.7$\pm$0.1 &96.7$\pm$0.1 \\
    SIG-VAE (K=1, J=1) \cite{hasanzadeh2019semi} &91.8$\pm$0.06 &93.0$\pm$0.08 &91.3$\pm$0.04 &92.4$\pm$0.04 &94.8$\pm$0.08 &95.2$\pm$0.06 \\
    SIG-VAE (K=15, J=20) \cite{hasanzadeh2019semi} &92.1$\pm$0.04 &93.2$\pm$0.06 &91.6$\pm$0.02 &92.7$\pm$0.03 &95.0$\pm$0.08 &95.4$\pm$0.04 \\
    SIG-VAE (K=150, J=2000) 
    \cite{hasanzadeh2019semi} &\underline{96.0}$\pm$0.04 &\underline{95.8}$\pm$0.06 &96.4$\pm$0.02 &\underline{96.3}$\pm$0.02 &\underline{97.0}$\pm$0.07 &\underline{97.2}$\pm$0.04\\
    \midrule
    WGCAE &95.0$\pm$0.02 &95.1$\pm$0.02 &\underline{96.5}$\pm$0.02 &\textbf{96.6}$\pm$0.02  &96.5$\pm$0.02 &96.7$\pm$0.02 \\
    WGAAE &\textbf{96.8}$\pm$0.02 &\textbf{96.6}$\pm$0.03 &\textbf{97.3}$\pm$0.03 &96.0$\pm$0.02 &\textbf{98.3}$\pm$0.04 &\textbf{97.7}$\pm$0.03 \\
    \bottomrule
    \end{tabular}}
\end{table*}

\begin{figure*}[t]
	\centering
	 \subfigure[$\Amat$]{\includegraphics[scale=0.36]{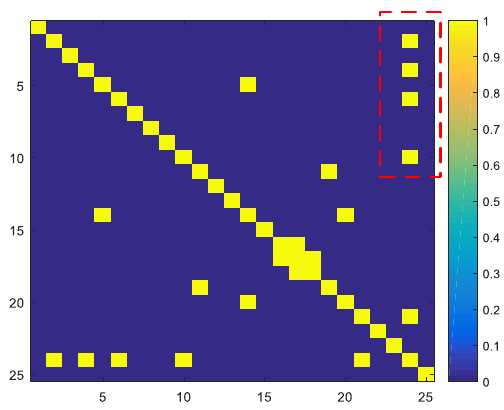}
		\label{fig:subgraph_A}}
	 \subfigure[$\Amat^{(1)}$]{\includegraphics[scale=0.36]{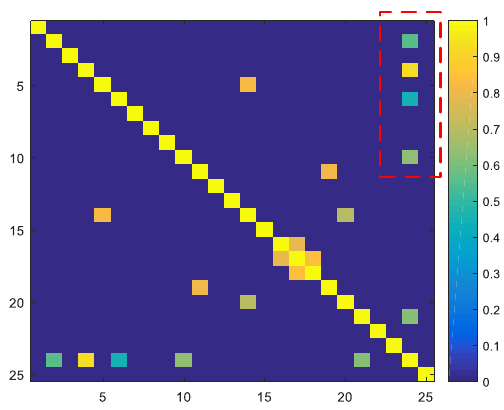}
		\label{fig:subgraph_B}}
	 \subfigure[$\Amat^{(2)}$]{\includegraphics[scale=0.36]{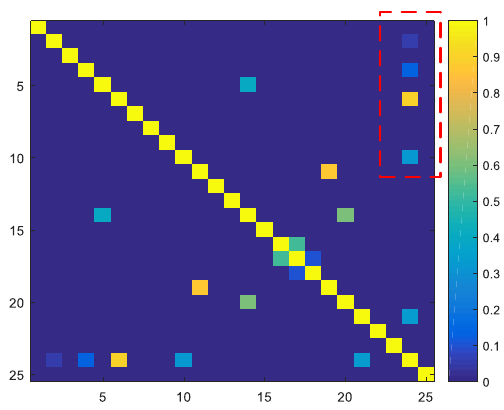}
		\label{fig:subgraph_C}}
	 \subfigure[$\Amat^{(3)}$]{\includegraphics[scale=0.36]{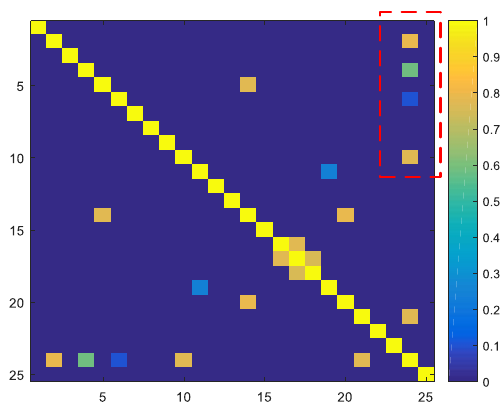}
		\label{fig:subgraph_D}} \\
	 \subfigure[$\Amat$]{\includegraphics[scale=0.36]{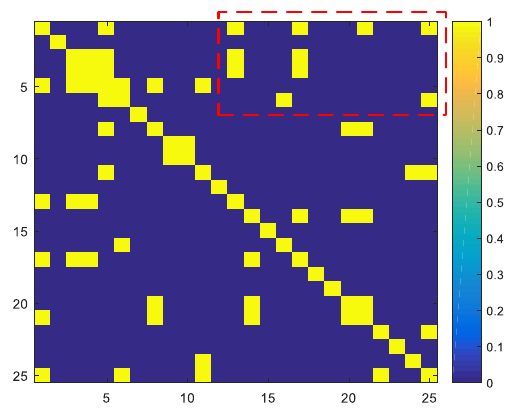}
		\label{fig:subgraph_E}}
	 \subfigure[$\Amat^{(1)}$]{\includegraphics[scale=0.36]{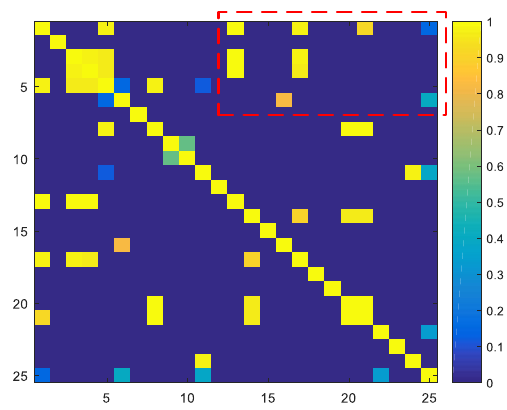}
		\label{fig:subgraph_F}}
	 \subfigure[$\Amat^{(2)}$]{\includegraphics[scale=0.36]{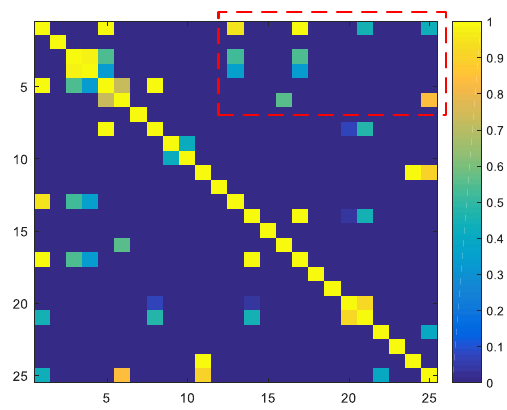}
		\label{fig:subgraph_G}}
	 \subfigure[$\Amat^{(3)}$]{\includegraphics[scale=0.36]{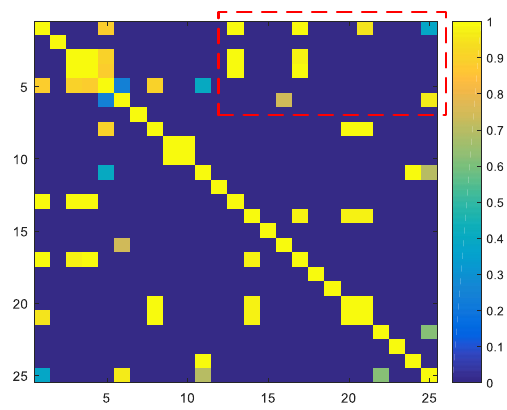}
		\label{fig:subgraph_H}}
	\caption{Visualization of part (randomly selected 25 nodes) of the hierarchical relationships learned by 3-layer WGCAEs on Cora (the first row) and Citeseer (the second row). The first column represents the observed adjacency matrix $\Amat$ and the second to fourth columns represent the learned adjacency matrices $\Amat^{(t)}$ from the layer 1 to 3, respectively. After normalization, a brighter point of each $\Amat^{(t)}$ indicates a stronger node relationship, and the red zone is highlighted for better demonstrations.}
	\label{fig:relationships}
\end{figure*}

\textbf{Datasets \& Preprocess}

Six widely used graph benchmarks are considered in the following quantitative experiments, including Cora, Citeseer, and Pubmed \cite{sen2008collective} for link prediction and node classification; Coil \cite{cai2010graph}, TREC \cite{CPGBN_ICML2019}, and R8 \cite{yao2019graph} for node clustering;
The detailed summary statistics of these benchmarks have been exhibited in Table \ref{tab:datasets}.

For link prediction and node classification, we use the node features and adjacency matrix provided by VGAE \cite{kipf2016variational}, and directly follow the preprocessing procedure in their released code.
With the same protocols as other methods, for link prediction, we train the model on an incomplete version of the network data, with 5\%/10\% of the citation links used for validation/test, and for node classification, the validation set is fixed to make a fair comparison.

For node clustering, following GNMF \cite{cai2010graph},  we manually construct a binary adjacency matrix via measuring the cosine similarity between documents and then thresholding the obtained matrix with Eq.~\eqref{Eq: adjacent matrix}.

\textbf{Model Settings}

For network structures, we construct a 3-layer WGCAE and WGAAE with the same network structure $K_1=K_2=K_3=C_k$, where $C_k$ is set as 16 for link prediction and the number of classes exhibited in Table~\ref{tab:datasets} for node clustering/classification. 
The head number of WGAAE is fixedly set as 4 and the hyper-paramerters of Bayesian attention mechanism are consistent those in \cite{fan2020Bayesian}. Other hyper-parameters in the generative network (decoder) are consistent to those settings in Table~\ref{tab:model_setting}. 

For optimization, the hyper-parameter $\beta$ in Eq.~\eqref{eq_ELBO} is selected by evaluating on the validation set of each benchmark, and the standard Adam \cite{kingma2014adam} with a learning rate of 1e-3 is utilized to optimize the loss function defined in Eq.~\eqref{eq_ELBO}. The details of model inference algorithms have been illustrated in Algorithm~\ref{algorithm_1} and \ref{algorithm_2}.

\begin{table*}[t]
    \centering
    \caption{Comparisons of node clustering performance.}
    \label{tab:graph_cluster}
    \scalebox{1}{
    \begin{tabular}{c|cc|cc|cc}
    \toprule
    \textbf{Method} &\multicolumn{2}{c|}{\textbf{Coil}} &\multicolumn{2}{c|}{\textbf{TREC}} &\multicolumn{2}{c}{\textbf{R8}} \\
    &ACC &NMI &ACC &NMI &ACC &NMI \\
    \midrule
    NMF \cite{cai2008non} &60.4$\pm$0.6 &72.6$\pm$0.5 &62.6$\pm$0.6 &45.5$\pm$0.7 & $54.6\pm$0.6 & 37.4$\pm$0.5 \\
    LDA \cite{blei2003latent}&59.4$\pm$0.5 &71.4$\pm$0.5 &60.9$\pm$0.7 &43.2$\pm$0.7 & $53.8\pm$0.6 & 36.9$\pm$0.6 \\
    PGBN \cite{zhou2016augmentable} &61.2$\pm$0.3 &73.4$\pm$0.4 &61.9$\pm$0.5 &44.1$\pm$0.4 & $54.7\pm$0.5 & 37.6$\pm$0.4\\
    \midrule
    GAE \cite{kipf2016variational} &65.8$\pm$0.4 & 77.0$\pm$0.4 &68.9$\pm$0.4 &52.8$\pm$0.3 &67.2$\pm$0.4 &44.5$\pm$0.4 \\
    VGAE \cite{kipf2016variational} &66.3$\pm$0.2 & 77.2$\pm$0.2 &69.0$\pm$0.3 &53.0$\pm$0.3 &67.4$\pm$0.3 &44.6$\pm$0.3 \\
    SIG-VAE \cite{hasanzadeh2019semi} &66.5$\pm$0.2 & 77.3$\pm$0.2 &69.2$\pm$0.4 &53.3$\pm$0.3 &67.5$\pm$0.3 &44.8$\pm$0.3 \\
    \midrule
    RTM \cite{chang2009relational} &70.7$\pm$0.6 &82.8$\pm$0.5 &71.5$\pm$0.7 &55.6$\pm$0.6 &70.4$\pm$0.7 &45.9$\pm$0.6\\
    GPFA &73.2$\pm$0.5 &84.6$\pm$0.4  &72.0$\pm$0.6 &56.1$\pm$0.6 &72.4$\pm$0.7 &46.8$\pm$0.5 \\
    GPGBN &73.6$\pm$0.5 &85.0$\pm$0.5  &72.3$\pm$0.5 &56.5$\pm$0.4 &73.3$\pm$0.5 &47.5$\pm$0.4\\
    GNMF \cite{cai2010graph} &78.7$\pm$2.5 &88.2$\pm$0.7 &72.5$\pm$1.8 &56.8$\pm$0.9 &73.8$\pm$1.4 &47.8$\pm$1.1 \\
    \midrule
    WGCAE  &\underline{83.3}$\pm$0.2 &\underline{89.5}$\pm$0.2 &\textbf{75.3}$\pm$0.3 &59.5$\pm$0.3 &\underline{78.2}$\pm$0.3 &\textbf{52.3}$\pm$0.3 \\
    WGAAE  &\textbf{87.0}$\pm$0.3 &\textbf{90.2}$\pm$0.3 
    &74.9$\pm$0.4 &\textbf{61.3}$\pm$0.3 &\textbf{79.2}$\pm$0.4 &\underline{51.6}$\pm$0.4 \\
    \bottomrule
    \end{tabular}}
\end{table*}

\textbf{Link Prediction}

Following the experimental settings in VGAE \cite{kipf2016variational}, we train the developed WGCAE/WGAAE on an incomplete version of the network data,  with $5\%$ and $10\%$ of the citation links used for validation and testing, respectively.
After training, the link prediction task is realized via link generation and we compare our models with other popular baselines, including DeepWalk (DW) \cite{perozzi2014deepwalk}, spectral clustering (SC) \cite{tang2011leveraging}, GAE and VGAE \cite{kipf2016variational},  SEAL \cite{zhang2018link},  G2G \cite{bojchevski2017deep}, $S$-VGAE \cite{davidson2018hyperspherical},  NF-VGAE, and SIG-VAE ($K$ and $J$ represents the sampling numbers of SIVI in every iteration)~\cite{hasanzadeh2019semi}.

Considering that the binary link prediction task is usually treated as a binary classification task, we evaluate the model performance with the average precision (AP) and area under the ROC curve (AUC), which is measured over 10 random training/testing splits.
As shown in the comparison results in Table~\ref{tab:link_predict}, we can find that the VGAE-based extensions in the second group generally outperform the vanilla GAE/VGAE, benefiting from providing more flexible posterior estimations rather than a Gaussian-reparameterized one.
Compared to these sophisticated VGAE-based methods, the developed WGCAE/WGAAE can explore hierarchical uncertainties and capture the multilevel relationships at multiple semantic layers, providing more expressive latent document representations to achieve better performance.
Moreover, the sparsity provided by the Weibull reparameterization and the introduction of the likelihood of node features in the loss function can effectively allevaite the developed WGCAE/WGAAE from overfitting and oversmoothing, leading to state-of-the-art or comparable link-prediction performance.
While SIG-VAE requires setting large $K$ and $J$ to achieve the state-of-the-art performance, which usually consumes an unaffordable memory footprint, the developed WGCAE/WGAAE requires much less memory to run and is more convenient to be implemented on personal platforms.

To intuitively illustrate the reason why extracting multilevel semantic relationships can contribute to improving the link-prediction performance, we visualize the link generation process learned by a 3-layer WGCAE on Cora and Citeseer, as shown in Fig.~\ref{fig:relationships}.
Specifically, after training, the developed WGCAE can ``\emph{divide}'' the observed adjacency matrix $\Amat$ into multiple adjacency matrices at different semantic layers $\{\Amat^{(t)}\}_{t=1}^T$, as discussed in Section \ref{subsec_model_properties}.
Then, for each graph dataset, we randomly select 25 nodes to exhibit the adjacency matrix $\Amat$ (see the first column of Fig. \ref{fig:relationships}), and the corresponding $\{\Amat^{(t)}\}_{t=1}^T$ (see columns 2-4 in Fig. \ref{fig:relationships}).
As the link structure fragment exhibited in Fig.~\ref{fig:subgraph_E}, which is highlighted with a red bounding box, we can find that the connection at the bottom right corner disappears at the first hidden layer but reoccurs at the second and third layers, which illustrates the effectiveness of exploring the relationships in multiple semantic layers rather than a single one.

\begin{table}[t]
    \centering
    \caption{Comparisons on node classification performance.}
    \label{tab:node_classification}
    \scalebox{1}{
    \begin{tabular}{c|ccc}
    \toprule
    \textbf{Method} &\textbf{Cora} &\textbf{Citeseer } &\textbf{Pubmed} \\
    \midrule
    ManiReg \cite{belkin2006manifold} &59.5 &60.1 &70.7  \\
    SemiEmb \cite{weston2012deep} &59.0 &59.6 &71.1  \\
    LP \cite{zhu2003semi} &68.0 &45.3 &63.0 \\
    DeepWalk \cite{perozzi2014deepwalk} &67.2 &43.2 &65.3 \\
    ICA \cite{lu2003link} &75.1 &69.1 &73.9 \\
    Planetoid \cite{yang2016revisiting} &75.7 &64.7 &77.2 \\
    GCN \cite{kipf2016semi} &81.5 &70.3 &79.0 \\
    GAT-16\textsuperscript{\ref{footnote}} \cite{velickovic2018graph} &82.3 &71.9 &78.7 \\
    GAT-64\textsuperscript{\ref{footnote}} \cite{velickovic2018graph} &83.0 &72.5 &79.0 \\
    DisenGCN\cite{DBLP:conf/icml/Ma0KW019}  &\underline{83.7} &\underline{73.4} &\textbf{80.5} \\
    SIG-VAE \cite{hasanzadeh2019semi} &79.7 &70.4 &{79.3} \\
    BAM-WF \cite{fan2020Bayesian} &83.5 &73.2  &78.5 \\
    \midrule
    WGCAE &82.0 &72.1 &79.1 \\
    WGAAE &\textbf{84.4} &\textbf{74.4} &\underline{80.2} \\
    \bottomrule
    \end{tabular}}
\end{table}

\textbf{Node Clustering}

For node clustering, we employ $K$-means on the concatenation of multilayer latent document representations $\{ \thetav _j^{(t)}\} _{t = 1}^T$, and then measure the clustering performance with the accuracy (AC) and normalized mutual information metric (NMI).
We compare the developed WGCAE/WGAAE with other related node clustering models, which can be roughly divided into three classes:
$i$) factorization based methods that model the generative process of node features without edge generation, including NMF \cite{cai2008non}, LDA \cite{blei2003latent}, and PGBN \cite{zhou2016augmentable};
$ii$) existing graph autoencoders, which have no node generation, including GAE, VGAE \cite{kipf2016variational}, and SIG-VAE \cite{hasanzadeh2019semi};
$iii$) methods that model both node and edge generations, like RTM \cite{chang2009relational} and GNMF~\cite{cai2010graph}.

As the node clustering results shown in Table~\ref{tab:graph_cluster}, the GAE-based approaches in the second group, which aggregate both the node and edge information via GNNs, generally outperform the factorization based methods in the first group that only consider the generation of node features.
However, limited by ignoring node generation, those GAEs that focus obsessively on the generation of the edges, generally achieve worse clustering performance than the methods in group three, indicating the importance of incorporating node generation in clustering tasks.
Among these graph-based methods in group three, attributed to more accurate posterior estimations, the proposed GPFA outperforms traditional RTMs, and the further performance improvement of GPGBN over GPFA, demonstrates that the richer semantics captured by a hierarchical probabilistic model can boost the clustering performance.
With a controllable weight to balance the importance of node and edge generations in the loss function, the developed WGCAE/WGAAE can further improve the clustering performance than these graph-based methods in group three.

\begin{table}[t]
 \centering
 \caption{Comparison on node classification performance  with different proportions of labeled nodes.}
 \label{tab_less_lables}
 {
 \scalebox{0.95}{
 \setlength{\tabcolsep}{2mm}{
 \begin{tabular}{c|cccc|cccc}
 \toprule
 \multirow{2}{*}{Method} 
 &\multicolumn{4}{c}{Cora}
 &\multicolumn{4}{c}{Citeseer} \\
 &80\%  &60\% &40\% &20\% &80\%  &60\% &40\% &20\% \\
 \midrule
 \multirow{1}*{GCN} \cite{kipf2016semi}
 &81.3 &80.6 &77.2 &\underline{73.1} 
 &72.1 &\underline{71.2} &70.2 &64.4 \\
 \multirow{1}*{GAT} \cite{velickovic2018graph} &\underline{83.4} 
 &\underline{80.6}  
 &77.2
 &72.0 
 &\underline{72.3} &70.6 &68.1 &61.3\\
 \midrule
 \multirow{1}*{WGCAE} 
 &80.0 &79.5 &\underline{77.4} &73.0
 &71.8 &70.9 &\underline{70.7} &\underline{66.1}\\
 \multirow{1}*{WGAAE} 
 &\textbf{84.2} &\textbf{82.7} &\textbf{80.6} &\textbf{76.0}
 &\textbf{73.2} &\textbf{72.8} &\textbf{72.2} &\textbf{70.9}\\
 \bottomrule
 \end{tabular}}}
 }
\end{table}

\textbf{Node Classification}

Owing to the VAE-liked model structure, the developed WGCAE/WGAAE is more convenient to plug in side information and can be flexibly extended for supervised learning, which remains a challenge for traditional probabilistic  RTMs.
In what follows, we evaluate the single-layer WGCAE/WGAAE with node classification task to make a further investigation.
Specifically, by introducing a categorical likelihood $p(y_j | \thetav_j)$ into Eq.~\eqref{eq_ELBO}, the  loss function of supervised WGCAE/WGAAE can be formulated as 
\begin{equation} \label{eq_node_classification_loss}
L_s = \sum_{j = 1}^N {\mathbb{E}}\left[ {\ln p({y_j}|\bm{\theta}_j^{(1)})}\right] + L,
\end{equation}
where $\{ y_j\}_{j=1}^{N}$ denotes the node labels.

Following SIG-VAE \cite{hasanzadeh2019semi}, we set the dimension of hidden layers as 16 for all methods to make a fair comparison, and the node classification results have been exhibited in Table~\ref{tab:node_classification}\footnote{Note the original GAT uses 64 hidden features (GAT-64), and we also report the results of GAT with 16 hidden features (GAT-16) to make a fair comparison. \label{footnote}}.
Despite not being trained specifically for node classification task, the developed WGCAE/WGAAE achieves comparable results on Cora, Citeseer, and Pubmed, which demonstrates the strong generalization properties of our models.
The advantages of WGCAE/WGAAE can be attributed to the following reasons:
$i$) jointly model the generative process of edges and node features can effectively alleviate 
overfitting, which is particularly serious in small graph datasets like Cora and Citeseer; 
$ii$) moving beyond discriminative node representations, the WGCAE/WGAAE can provide hierarchical stochastic
latent representations with the Weibull distribution, which is more suitable to approximate sparse and skewed latent document representations.

To further investigate the advantages of incorporating the  reconstruction objective into the training loss of node classification task, as shown in Eq.~\eqref{eq_node_classification_loss}, we compare the developed WGCAE/WGAAE with other baselines with different proportions of labeled nodes (80$\%$, 60$\%$, 40$\%$, 20$\%$). 
As shown by the classification results in Table~\ref{tab_less_lables},  we can find that the gain of incorporating graph reconstruction into model training becomes more and more evident as the number of labels decreases.
Thus, in practice, the developed WGCAE/WGAAE could be a more robust choice under the situation of sparsely labeled nodes, which is a common feature to have for real-world graphs.



\subsection{Ablation Study}

In this part, we carefully design a series of experiments for ablation study to investigate the property of our developed methods.

\textbf{Balance between the Node and Edge Generations}

\begin{figure}[t]
	 \subfigure[Node Clustering]{\includegraphics[scale=0.45]{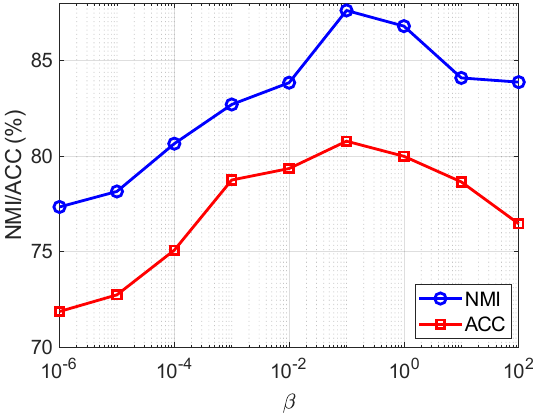}
		\label{beta_cluster}}
	 \subfigure[Link Prediction]{\includegraphics[scale=0.45]{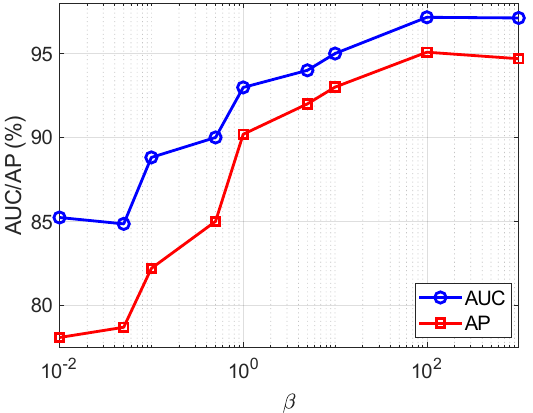}
		\label{beta_link}}
	\caption{The effect of trade-off hyper-parameter $\beta$ on (a) node clustering task and (b) link prediction task.}
	\label{fig: beta}
\end{figure}

As discussed in Section~\ref{sec_wgcae}, for GPFA/GPGBN trained with Gibbs sampling, there remains a challenge to control the trade-off between the likelihoods of node features and adjacency matrix for different tasks, which potentially limit the model capabilities.
To this end, WGCAE/WGAAE introduces a controllable hyperparameter $\beta$ into the loss function Eq.~\eqref{eq_ELBO} to balance the focuses on nodes and edges, which allows an 
adjustable
treatment of node features $\Xmat$ and adjacency matrix~$\Amat$.
To evaluate the effectiveness of the trade-off hyperparameter~$\beta$, we perform experiments with WGCAE on Coil dataset under the different value settings of $\beta$, and investigate the corresponding influence on the model performance of both node clustering and link prediction tasks.
As shown in Fig.~\ref{fig: beta}, we can find that the ideal choice of $\beta$ to achieve the best performance of node clustering and link prediction is different, where the former is around $\beta=0.1$ and the latter could be $\beta=100$.
This phenomenon verifies the different contributions of node and edge generations in various graph analytic tasks, and a larger $\beta$ in Eq.~\eqref{eq_ELBO} will control the model to pay more attention to the reconstruction of adjacency matrix rather than node features.
Besides, it potentially explains the reason why traditional GAEs can achieve promising performance on link prediction but poor performance on node clustering, since they only consider the generation of edges.

\textbf{Efficiency of Scalable Inference}

\begin{figure}[t]
	 \subfigure[Cora]{\includegraphics[scale=0.45]{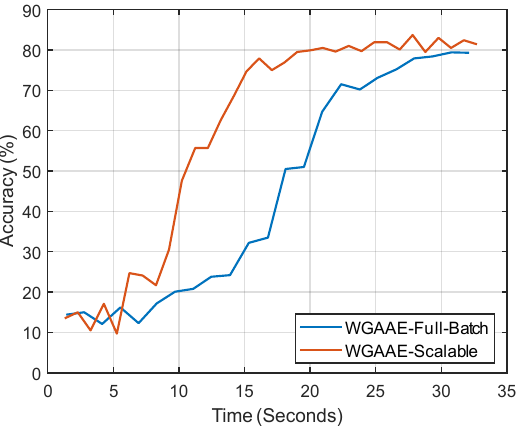}
		\label{efficiency cora}}
	\quad
	 \subfigure[Pubmed]{\includegraphics[scale=0.45]{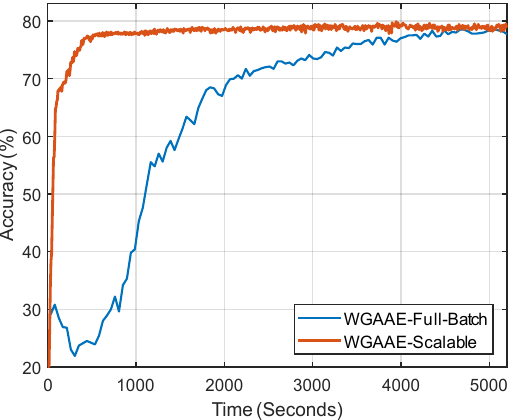}
		\label{efficiency citeseer}}
	\caption{Efficiency comparisons of different training algorithms on the node classification task on (a) Cora and (b) Pubmed datasets.}
	\label{fig: efficiency}
\end{figure}

To demonstrate the efficiency of the developed scalable training algorithm in Section~\ref{scalable}, under the same network structure and hyperparameter settings, we train the WGAAE with two different training algorithms to perform the node classification task, where the one is full-batch (non-scalable) training algorithm in Algorithm~\ref{algorithm_1} and the other is the scalable version based on minibatch estimation, as described in Algorithm~\ref{algorithm_2}.
Then we plot the curves of classification accuracy versus training time on Cora and Pubmed datasets, respectively, as shown in Fig.~\ref{fig: efficiency}, where the minibatch size $N_s$ is set as 100.
From the results, we can find that the developed scalable inference algorithm can significantly speed up the training procedure for model convergence, under the premise of almost no loss of model performance.

Additionally, we also evaluate the developed WGAAE equipped with the scalable inference algorithm, as shown in Algorithm \ref{algorithm_2}, on the recent popular OGB Large-Scale Challenge (OGB-LSC) \cite{DBLP:conf/nips/HuFRNDL21}. 
Following the official experimental instruction provided by OGB-LSC, we make a comparison with a series of scalable baselines on MAG240M dataset under the homogeneous settings\footnote{Note that the results of heterogeneous scalable baselines are omitted here for a fair performance comparison. \label{footnote2}}, where MAG240M is an extremely large-scale academic graph extracted Microsoft Academic Graph (MAG) \cite{wang2020microsoft} and the task is to predict the subject areas of papers.
We emphasize that the network structure and hyper-parameter settings of WGAAE here are  consistent with those in the previous experiments for convenience, which are not deliberately selected based on the best validation performance like other reported baselines, and the novel-level prediction performance of WGAAE can still outperform other scalable baselines as shown in Table~\ref{tab:node_prediction}, demonstrating the effectiveness of our methods for handling with extremely large-scale graphs.

\begin{table}[t]
    \centering
    \caption{Comparisons on node-level prediction performance on OGB-LRC MAG240M dataset.}
    \label{tab:node_prediction}
    \scalebox{1}{
    \begin{tabular}{c|cc}
    \toprule
    \textbf{Method} &\textbf{Validation} &\textbf{Test } \\
    \midrule
    MLP &52.67 &52.73 \\
    LabelProp &58.44 &56.29 \\
    SGC \cite{wu2019simplifying} &65.82 &65.29  \\
    SIGN \cite{frasca2020sign} &66.64 &66.09  \\
    MLP+C\&S \cite{huang2020combining} &66.98 &66.18 \\
    GraphSAGE (NS) \cite{hamilton2017inductive} &66.79 &66.28 \\
    GAT (NS) \cite{velickovic2018graph} &\underline{67.15} &\underline{66.80} \\
    \midrule
    WGAAE &\textbf{67.23} &\textbf{66.91} \\
    \bottomrule
    \end{tabular}}
\end{table}




\section{Conclusion}

To alleviate the flexibility issues when extending VGAEs for modeling document relational networks (DRNs), in this paper, we develop a non-Gaussian VGAE-liked model with multiple stochastic hidden layers, which incorporates a novel deep RTM as its generative network (decoder).
Specifically, we first develop a graph Poisson factor analysis (GPFA) to model the generation of interconnected documents, which provides analytic conditional posteriors for all model parameters without introducing sophisticated approximate assumptions like traditional RTMs.
Then we extend the single-layer GPFA to a multilayer graph Poisson gamma belief network (GPGBN), which is the first unsupervised deep (hierarchical) RTM to explore hierarchical latent document representations and semantic relationships from DRNs.
To achieve fast in out-of-sample prediction and flexible to plug in side information, we combine GPGBN (decoder) with two different forms of Weibull-based graph variational inference networks (encoder), one based on the vanilla GCN and the other based on the Bayesian attention mechanism, resulting in two variants of Weibull graph autoencoder (WGAE).
For efficient end-to-end model training, we propose both full-batch (non-scalable) and scalable versions of hybrid model inference algorithms for the developed WGAEs, where the latter is scalable to large-scale graphs.
Experimental results demonstrate that our models can extract high-quality hierarchical latent document representations, leading to improved performance over baselines on a series of graph analytic tasks, and also provide a potential solution to exploring explainable link structure prediction at multiple semantic levels.



\section*{Acknowledgment}
This work was supported in part by the National Natural Science Foundation of China under Grant U21B2006; in part by Shaanxi Youth Innovation Team Project; in part by the 111 Project under Grant B18039; in part by the Fundamental Research Funds for the Central Universities QTZX22160.

\bibliographystyle{IEEEtran}
\bibliography{chaos_ref}

\newpage
\onecolumn
\appendix
\section{Link Functions for Different Types of Observations}\label{ap_link_function}

In this paper, for decoder, we first develop GPFA, which is a shallow topic relational model (RTM) equipped with analytic conditional posteriors,  and further extend it to GPGBN,  a hierarchical (deep) generalization.
For brevity, in the main manuscript, we only give a detailed discussion on the binary type of adjacency matrix $\Amat$ and a short introduction on how to build other types of $\Amat$.
Thus, in this section, we will provide more detailed discussion on the continuous nonnegative and count-valued types of $\Amat$, which are also usually used in practice besides the binary type.
Considering the likelihood of $\Amat$ is similar in GPFA and GPGBN, in the following, we take the observed adjacency matrix $\Amat$ in GPFA as an example to discuss.
For completeness, we first discuss the binary  $\Amat$ and then extend it to the continuous nonnegative and count-valued types.

{\bf{Binary adjacency matrix: }}
If the edge $a_{ij}$ is binary to model whether two documents are associated, we can use a Bernoulli-Poisson (BerPo) link \cite{zhou2015infinite} expressed as
\begin{equation}
{a_{ij}} = 1({m_{ij}} > 0),{m_{ij}}\sim \mbox{Pois}(\sum\nolimits_{k = 1}^{{K_1}} {{u_k}{\theta _{ik}}} {\theta _{jk}})
\end{equation}
where $a_{ij}=1$ if ${m_{ij}} \ge 1$ and $a_{ij}=0$ if ${m_{ij}} = 0$.
After $m_{ij}$ is marginalized out, we can obtain a Bernoulli random variable as ${a_{ij}}\sim \mbox{Bern}(1 - \exp ( - \sum\nolimits_{k = 1}^{{K_1}} {{u_k}{\theta _{ik}}} {\theta _{jk}}))$.
The conditional posterior of the latent count ${m_{ij}}$ can be expressed as
\begin{equation}
({m_{ij}}\given {a_{ij}},\bm{u},\bm{\theta _i},\bm{\theta _i})\sim {a_{ij}} \cdot \mbox{Pois}_{+}(\sum\nolimits_{k = 1}^{{K_1}} {{u_k}{\theta _{ik}}} {\theta _{jk}}),
\end{equation}
which can be simulated with a rejection sampler as described in \cite{zhou2015infinite}.

{\bf{Continuous nonnegative adjacency matrix: }}
If the edge $a_{ij}$ is a continuous nonnegative value, like the cosine similarity between two documents, we
can use gamma-Poisson link \cite{zhou2016augmentable} expressed as
\begin{equation}\label{gam_poisson}
{a_{ij}}\sim \mbox{Gam}({m_{ij}},1/c),{m_{ij}}\sim \mbox{Pois}(\sum\nolimits_{k = 1}^{{K_1}} {{u_k}{\theta _{ik}}} {\theta _{jk}})
\end{equation}
whose distribution has a point mass at $a_{ij}=0$ and is continue for $a_{ij}>0$. Further the latent count ${m_{ij}}=0$ if and only if $a_{ij}=0$, and ${m_{ij}}$ is  a positive integer drawn from a truncated Bessel distribution if $a_{ij}>0$.

{\bf{Count-valued adjacency matrix: }}
If the edge $a_{ij}$ is a count value, such as indicating the frequency of citation, we can directly factor the discrete value $a_{ij}$ under poisson likelihood as
\begin{equation}
{a_{ij}}\sim \mbox{Pois}(\sum\nolimits_{k = 1}^{{K_1}} {{u_k}{\theta _{ik}}} {\theta _{jk}})
\end{equation}

To a conclusion, our models can be generalized to different types of networks with nodes and edges
expressed by count, binary, or positive values via making full use of different link functions.

\section{Derivation for GPGBN}\label{ap_gibbs}

Here we describe the detailed derivation for graph Poisson gamma belief network (GPGBN) with $T$ hidden layers, expressed as
\begin{align}\label{GPGBN}
&\bm{x}_j^{(1)} \sim \mbox{Pois}({\bm{\Phi} ^{(1)}}\bm{\theta} _j^{(1)}),
\left\{ {\bm{\theta} _j^{(t)}\sim \mbox{Gam}({\bm{\Phi} ^{(t + 1)}}\bm{\theta} _j^{(t + 1)},1/c_j^{(t + 1)})} \right\}_{t = 1}^{T - 1},
\bm{\theta} _j^{(T)} \sim \mbox{Gam}(\bm{\gamma} ,1/c_j^{(T + 1)}), \notag\\
&{a_{ij}}= 1({\delta _{ij}} \textgreater 1),\quad {\delta _{ij}} = \sum\nolimits_{t = 1}^T {m_{ij}^{(t)}}, \quad
\left\{ {m_{ij}^{(t)}\sim \mbox{Pois}(\sum\nolimits_{k = 1}^{{K_t}} {u_k^{(t)}\theta _{ik}^{(t)}\theta _{jk}^{(t)}} )} \right\}_{t = 1}^T,
\end{align}

\subsection{Property}

During the inference procedure, we adopt the variable augmentation and marginalization techniques of PGBN \cite{zhou2016augmentable}, exploiting the following  properties of the Poisson, gamma, and related distributions:

{\bf{Property 1 (P1): }}
If ${x. = \sum\limits_{n = 1}^N {{x_n}} }$, where ${{\mbox{ }}{x_n} \sim {\mathop{\mbox{Poisson}}\nolimits} \left( {{\theta _n}} \right)}$ are independent Poisson-distributed random variables, then we have ${\left( {{x_1}, \ldots ,{x_N}} \right) \sim {\mathop{\mbox{Multinomial}}\nolimits} \left( {x,\frac{{{\theta _1}}}{{\sum\nolimits_{n = 1}^N {{\theta _n}} }}, \ldots ,\frac{{{\theta _N}}}{{\sum\nolimits_{n = 1}^N {{\theta _n}} }}} \right)}$ and ${{x_ \cdot } \sim {\mathop{\mbox{Poisson}}\nolimits} \left( {\sum\nolimits_{n = 1}^N {{\theta _n}} } \right)}$. 

{\bf{Property 2 (P2): }}
If ${x \sim {\mathop{\mbox{Poisson}}\nolimits} (c\theta )}$, where $c$ is a constant and ${\theta \sim {\mathop{\rm Gam}\nolimits} (a,1/b)}$, then we can marginalize out $\theta$ and obtain
${{\mbox{ }}x \sim {\mathop{\rm NB}}\left( {a,\frac{c}{{b + c}}} \right)}$, satisfying a negative binomial (NB) distribution.

{\bf{Property 3 (P3): }}
If ${{x} \sim {\mathop{\rm NB}\nolimits} (a, p)}$ and ${l \sim {\mathop{\rm CRT}\nolimits} (x,a)}$ is a Chinese restaurant table-distributed random variable, then $x$ and $l$ are equivalently jointly distributed as ${x \sim {\mathop{\rm SumLog}\nolimits} (l,p)}$ and ${l \sim {\mathop{\mbox{Poisson}}\nolimits} (-\ln(1-p)a )}$ \cite{Zhou2015Negative}.

\subsection{Gibbs Sampling}

Making full use of these properties, we can obtain the following Poisson likelihoods for each hidden layer of GPGBN, formulated as:
\begin{align}
x_{{k_{t - 1}}j}^{(t)} &\sim \mbox{Pois}( - \ln (1 - p_j^{(t)})\sum\nolimits_{k = 1}^{{K_t}} {\phi _{{k_{t - 1}}k}^{(t)}\theta _{jk}^{(t)}} ),\label{data_likelihood}\\
m_{ij}^{(t)} &\sim \mbox{Pois}(\sum\nolimits_{k = 1}^{{K_t}} {u_{k}^{(t)} \theta _{ik}^{(t)}\theta _{jk}^{(t)}} ),
\end{align}
where the augmented node feature vector $\bm{x}_j^{(t)} \in \mathbb{R}_ + ^{{K_{t - 1}}}$ and adjacency matrix ${\bm{M}^{(t)}} \in {\mathbb{Z}^{N \times N}}$ can be regarded as the specific observations at layer $t$. Thus, for each hidden layer of GPGBN, the analytic conditional posteriors for model parameters can be formulated as follows:

{\bf{Sampling loading factor matirx $\bm{\Phi}^{(t)}$: }} Utilizing the simplex constraint on each column of ${\bm{\Phi} ^{(t)}} \in {\mathbb{R}_{+}^{{K_{t - 1}} \times {K_t}}}$, we can marginalize out $\bm{\theta}_j$ in (\ref{data_likelihood}) and have
\begin{equation}
(x_{1j{k_t}}^{(t)},...,x_{{K_{t - 1}}j{k_t}}^{(t)}|x_{ \cdot j{k_t}}^{(t)})\sim \mbox{Multi}(x_{ \cdot j{k_t}}^{(t)}\given \bm{\phi} _{{k_t}}^{(t)}).
\end{equation}
Therefore, via the Dirichlet-multinomial conjugacy, the posterior of $\bm{\phi}^{(t)}_{k_t}$ can be formulated as
\begin{equation}
(\bm{\phi} _{{k_t}}^{(t)}\given - )\sim \mbox{Dir}(x_{1 \cdot {k_t}}^{(t)} + {\eta ^{(t)}},...,x_{{K_{t - 1}} \cdot {k_t}}^{(t)} + {\eta ^{(t)}}).
\end{equation}

{\bf{Sampling topic proportions $\bm{\theta} _j^{(t)}$: }} Benefit from jointly modeling node features and link structure under poisson likelihood, we have
\begin{equation}
\begin{array}{r}
({\theta^{(t)}_{jk_t}}\given - ) \sim  \mbox{Gam}\big({x^{(t)}_{ \cdot jk_t}} + {\bm{\phi} ^{(t+1)}_{k_t:}}\bm{\theta} _j^{(t+1)} \ns+\ns \sum\limits_{i \ne j} {{m^{(t)}_{ijk_t}}}, {[-\ln(1-p^{(t)}_j) + {c^{(t+1)}_j} + {u^{(t)}_{k_t}}\sum\limits_{i \ne j} {{\theta^{(t)}_{ik_t}}} ]^{ - 1}} \big).
\label{eq_theta_posterior}
\end{array}
\end{equation}
where ${x_{ \cdot jk_t}^{(t)}}$ and ${{m_{ijk_t}^{(t)}}}$ are latent count variables that are independently sampled from the corresponding node feature and relative edges at layer $t$, respectively.

{\bf{Sampling scale parameter $c^{(t)}_j$ and $p^{(t)}_j$: }} To construct a hierarchical generative model, we introduce $c_j^{(t)}\sim \mbox{Gam}({e_0},1/{f_0})$ and the corresponding posterior of $c^{(t)}_j$ can be formulated as
\begin{equation}
(c_j^{(t)}\given - )\sim \mbox{Gam}(\theta _{j\cdot}^{(t)}  + {e_0},1/[{f_{\rm{0}}}{\rm{ + }}\theta _{ j\cdot}^{(t - 1)}]).
\end{equation}
Referring the Lemma \textbf{1} of \cite{zhou2015poisson}, ${\{ p_j^{(t)}\} _{t \ge 2}}$ can be calculated with
\begin{equation}
p_j^{(t + 1)}:= - \ln (1 - p_j^{(t)})/[c_j^{(t + 1)} - \ln (1 - p_j^{(t)})],
\end{equation}
specifically defining $p_j^{(1)}: = 1 - {e^{ - 1}}$.

{\bf{Sampling $\bm{u}^{(t)}$: }} With the prior $u_{{k_t}}^{(t)}\sim \mbox{Gam}({\alpha _0},1 / {\beta _0})$, via the gamma-Poisson conjugacy, we have
\begin{align}
(u_{{k_t}}^{(t)}\given - )  \sim & \mbox{Gam}(\prod\nolimits_{i = 1}^N {\prod\nolimits_{j = 1}^{i - 1} {m_{ij{k_t}}^{(t)}} }  + {\alpha _{{0}}}, {[{\beta _{{0}}} + \prod\nolimits_{i = 1}^N {\prod\nolimits_{j = 1}^{i - 1} {\theta _{i{k_t}}^{(t)}\theta _{j{k_t}}^{(t)}} } ]^{ - 1}}).
\end{align}

\subsection{Time Complexity Analysis}

Compared to the basic PGBN, the additional time cost for GPGBN is mainly in the procedure of graph augmentation, which can be formulated as
\begin{align}
&(m_{ij}^{(1)},...,m_{ij}^{(T)}\given {m_{ij}}) \sim \mathop{\mbox{Multi}}({m_{ij}};\frac{{\sum\nolimits_{{k_1} = 1}^{{K_1}} {u_{{k_t}}^{(t)}\theta _{i{k_1}}^{(t)}\theta _{j{k_1}}^{(t)}} }}{{\sum\nolimits_{t = 1}^T {\sum\nolimits_{{k_t} = 1}^{{K_t}} {u_{{k_t}}^{(t)}\theta _{i{k_t}}^{(t)}\theta _{j{k_t}}^{(t)}} } }},...,\frac{{\sum\nolimits_{{k_T} = 1}^{{K_T}} {u_{{k_T}}^{(t)}\theta _{i{k_T}}^{(t)}\theta _{j{k_T}}^{(t)}} }}{{\sum\nolimits_{t = 1}^T {\sum\nolimits_{{k_t} = 1}^{{K_t}} {u_{{k_t}}^{(t)}\theta _{i{k_t}}^{(t)}\theta _{j{k_t}}^{(t)}} } }}),\\
&(m_{ij1}^{(t)},...,m_{ij{K_t}}^{(t)}\given m_{ij}^{(t)}) \sim \mathop{\mbox{Multi}}(m_{ij}^{(t)};\frac{{u_1^{(t)}\theta _{i1}^{(t)}\theta _{j1}^{(t)}}}{{\sum\nolimits_{{k_t} = 1}^{{K_t}} {u_{{k_t}}^{(t)}\theta _{i{k_t}}^{(t)}\theta _{j{k_t}}^{(t)}} }},...,\frac{{u_{{K_t}}^{(t)}\theta _{i{K_t}}^{(t)}\theta _{j{K_t}}^{(t)}}}{{\sum\nolimits_{{k_t} = 1}^{{K_t}} {u_{{k_t}}^{(t)}\theta _{i{k_t}}^{(t)}\theta _{j{k_t}}^{(t)}} }})
\end{align}
where the former indicates the multi-layer augmentation from ${m_{ij}}$ and the latter denotes the augmentation at layer $t$.
Here we note that there is no need to augment the whole adjacency matrix $\Amat$, but only the positive elements (edges)
$\{ {a_{ij}} > 0\} _{i = 1,j = 1}^{N,N}$ in $\Amat$.
Thus the computational benefit is significant since the computational complexity is approximately linear to the number of $\{ {a_{ij}} > 0\} _{i = 1,j = 1}^{N,N}$, denoted as $S_a$, in the observed adjacency matrix $\Amat$.
Compared to directly handling with the whole matrix like traditional RTMs \cite{chang2010hierarchical}, this benefit is especially pertinent in many real-world network data where $S_a$ is
significantly smaller than $N^2$.
Moreover, the augmentation operations for node features $\Xmat$ or link structure $\Amat$ can be both processed with parallel Gibbs sampling \cite{CPGBN_ICML2019}, which can be further accelerated with GPU.

\end{document}